\documentclass{article}

\usepackage[numbers]{natbib}
\usepackage[preprint]{neurips_2025}

\usepackage[utf8]{inputenc} 
\usepackage[T1]{fontenc}    
\usepackage{hyperref}       
\usepackage{url}            
\usepackage{booktabs}       
\usepackage{tablefootnote}  
\usepackage{multirow}        
\usepackage{amsfonts}       
\usepackage{nicefrac}       
\usepackage{microtype}      
\usepackage{amsmath}
\usepackage{graphicx}
\usepackage{wrapfig}
\usepackage{makecell}

\usepackage[ruled,vlined]{algorithm2e}
\usepackage[dvipsnames]{xcolor}       
\usepackage{etoolbox}
\usepackage{caption}
\AtBeginEnvironment{figure}{\footnotesize}
\AtBeginEnvironment{table}{\footnotesize}
\usepackage{cleveref}      
\usepackage[misc]{ifsym} 
\definecolor{myblue}{HTML}{0064E0}
\definecolor{myred}{HTML}{DC6F7B}
\hypersetup{
    colorlinks,
    linkcolor=myblue,   
    citecolor=myblue,    
    filecolor=Mulberry,       
    urlcolor=RoyalBlue,       
}
\usepackage{pifont}  
\newcommand{\cmark}{\ding{51}} 

\title{Rethinking Agent Design: From Top-Down Workflows to Bottom-Up Skill Evolution}
\author{
Jiawei Du\textsuperscript{1,2}
Jinlong Wu\textsuperscript{1,2,3} 
Yuzheng Chen\textsuperscript{1,2,3} 
Yucheng Hu\textsuperscript{4} 
Bing Li\textsuperscript{5} 
Joey Tianyi Zhou\textsuperscript{1,2\,\Letter}\\
{\small \textsuperscript{1} Centre for Frontier AI Research (CFAR), Agency for Science, Technology and Research (A*STAR), Singapore}\\
 {\small \textsuperscript{2} Institute of High Performance Computing, Agency for Science, Technology and Research (A*STAR), Singapore}\\
\textsuperscript{3}{\small National University of Singapore, Singapore}  \textsuperscript{4}{\small Tsinghua University, Beijing, China}\\
\textsuperscript{5}{\small University of Electronic Science and Technology of China, Chengdu}  \\
}
\begin{document}
\maketitle
\begin{abstract}
\vspace{-1em}
Most LLM-based agent frameworks adopt a top-down philosophy: humans decompose tasks, define workflows, and assign agents to execute each step. While effective on benchmark-style tasks, such systems rely on designer updates and overlook agents’ potential to learn from experience. Recently, Silver and Sutton~\cite{eraofexp} envision a shift into a new era, where agents could progress from a stream of experiences. In this paper, we instantiate this vision of experience-driven learning by introducing a bottom-up agent paradigm that mirrors the human learning process. Agents acquire competence through a trial-and-reasoning mechanism—exploring, reflecting on outcomes, and abstracting skills over time. Once acquired, skills can be rapidly shared and extended, enabling continual evolution rather than static replication. As more agents are deployed, their diverse experiences accelerate this collective process, making bottom-up design especially suited for open-ended environments. We evaluate this paradigm in Slay the Spire and Civilization V, where agents perceive through raw visual inputs and act via mouse outputs, the same as human players. Using a unified, game-agnostic codebase without any game-specific prompts or privileged APIs, our bottom-up agents acquire skills entirely through autonomous interaction, demonstrating the potential of the bottom-up paradigm in complex, real-world environments. Our code is available at \href{https://github.com/AngusDujw/Bottom-Up-Agent}{https://github.com/AngusDujw/Bottom-Up-Agent}.
\end{abstract}

\section{Introduction}
\begin{wrapfigure}{R}{0.5\textwidth}
 \centering
 \includegraphics[width =0.5\textwidth]{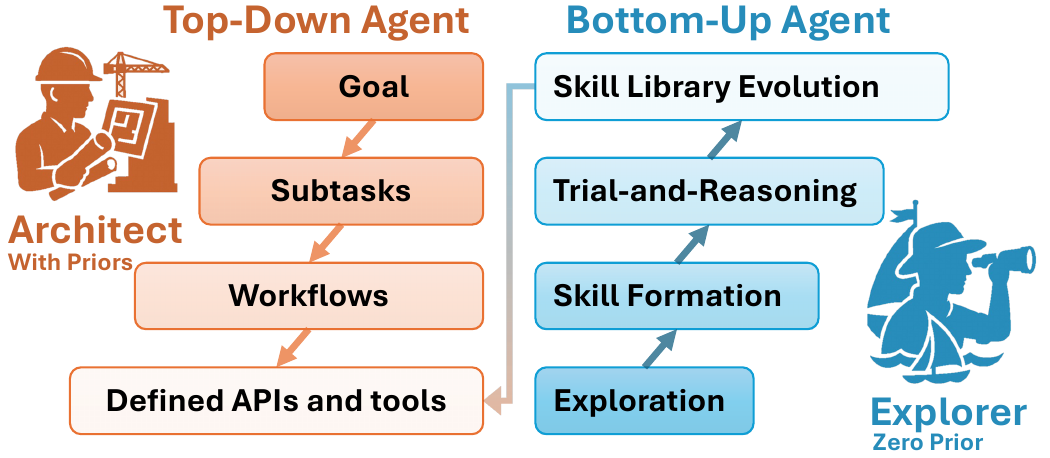}
 \caption{ \textbf{Two paradigms of agent design.} Most existing agent frameworks can be categorized as \textbf{Top-down agents}, which rely on pre-engineered architectures: they begin with high-level goals, decompose them into subtasks, and execute workflows using task-specific APIs and tools. In contrast, we propose \textbf{Bottom-Up agents} to function as explorers: starting from zero prior knowledge, they gradually acquire skills through trial-and-reasoning, evolving autonomously via implicit reward inferred from environmental change.}
 \label{fig:intro}
 \vspace{-1em}
\end{wrapfigure}
\label{sec:intro}
The advance in Large Language Models (LLMs) has propelled agentic AI systems toward increasingly complex task-solving capabilities~\cite{gpt4,deepseek,claude37,janus,uitars}. Most existing agent frameworks~\cite{metagpt,autogpt,chatdev,generative_agent,voyager} follow a top-down design philosophy: given a high-level goal, humans decompose it into subtasks, design a static or dynamic workflow, and assign agents to each node. To maintain alignment with intended goals, a memory module is often included to correct deviations. These systems~\cite{react,reflexion,metagpt,autogpt} are optimized for execution and correction, which makes them highly effective on well-defined, benchmark-style tasks~\cite{bench1,bench2,bench3,bench4,osworld}, but limiting their scalability in open-ended, real-world environments.

This limitation arises from the top-down paradigm’s roots in traditional software engineering, which impose three key constraints: (i) Staticness: Agents are deployed as identical replicas of a central prototype~\cite{chatdev,metagpt,manus,mas_fail}, with improvements pushed manually rather than learned adaptively. (ii) Prior Dependency: The top-down agents rely on predefined APIs~\cite{vpt,voyager,nature_mine,dreamerv3,mas_fail}, workflows, and task-specific prompts. In open-ended environments where such structures are unavailable, they struggle to initiate meaningful behavior. (iii) Token Utilization: A large portion of LLM tokens~\cite{token_eff1,token_eff2} are consumed enforcing predefined workflows. These tokens could instead be used for reasoning over lived experience.


These constraints confine agent improvement to manual designer intervention, inhibiting autonomous evolution during deployment. Fortunately, the recently unlocked reasoning capabilities of LLMs enable a new generation of agents, which acquire skills not by mimicking expert trajectories, but by continuously learning from streams of grounded interactions. These agents, as envisioned by ~\citet{eraofexp}, mark a shift into the era of experience, in contrast to the prior era dominated by curated human data~\cite{eraofexp}.

Building on this vision, we introduce a bottom-up agent paradigm that instantiates the principle of learning from experience. This paradigm leverages the reasoning capabilities of LLMs~\cite{reasoning1,reasoning2,deepseek} to enable agents to acquire skills autonomously during deployment. In contrast to the execution-and-correction loop typical of top-down systems, the bottom-up paradigm prioritizes exploration and reasoning. We illustrate the distinction between top-down and bottom-up paradigms in \autoref{fig:intro}.

\begin{figure}[h] 
\vspace{-1.em}
	 \centering
	 \includegraphics[width = 1.0\linewidth]{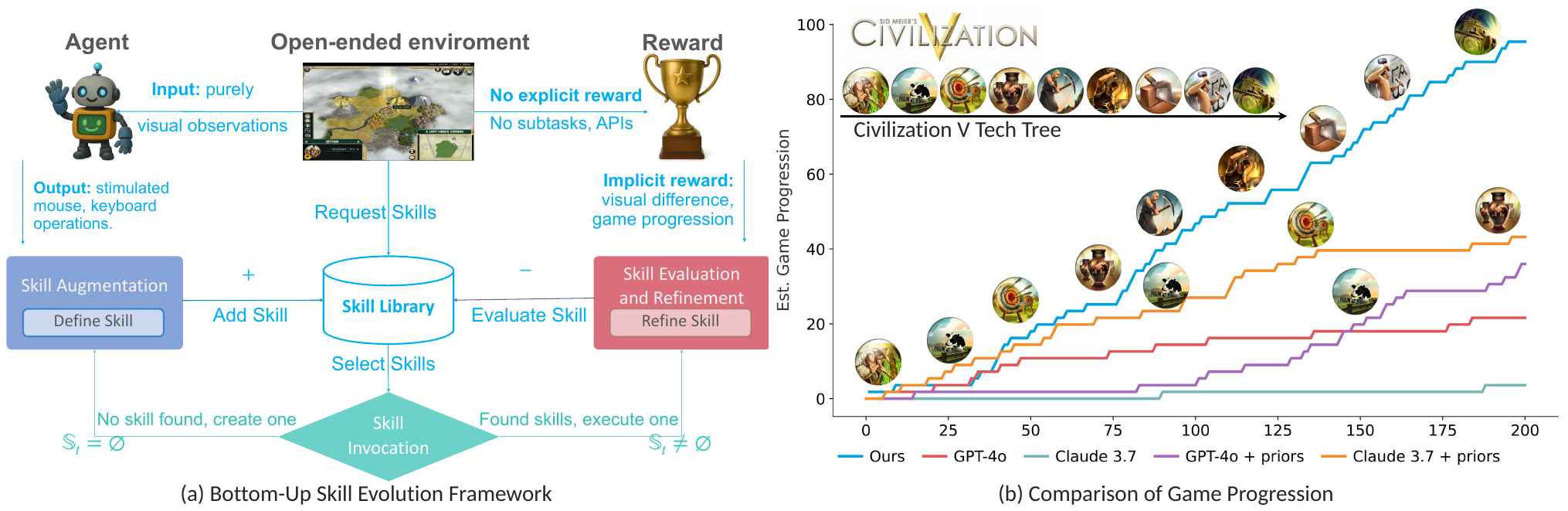}
	 \caption{\textbf{Left:} The bottom-up agent operates solely on raw visual input and simulates low-level mouse and keyboard actions. Without explicit rewards, it learns and refines skills based on implicit signals like visual changes or game progression. \textbf{Right:} Game progression measured by Civilization V’s tech tree and visual changes. Our bottom-up agent (blue) outperforms all baselines, including those with task-related priors. }
     \label{fig:pipeline}
\vspace{-1.em}
\end{figure}

The bottom-up paradigm mirrors the human learning process, emphasizing experiential learning through autonomous exploration. Like humans, agents acquire competence through trial-and-reasoning interaction: they autonomously execute skills, reason about their consequences, and iteratively refine their skill libraries. Skill rewards are implicitly inferred from environmental feedback, which consists of visual changes, game progression, or numeric indicators—without requiring explicit supervision.  To prevent redundant exploration and accelerate collective progress, the bottom-up paradigm also draws inspiration from the diffusion of human technologies: once a superior skill is discovered by any agent, it can be rapidly disseminated through a cloud-based knowledge-sharing mechanism. This enables all agents to immediately access and build upon the most effective strategies identified in the field, fostering a self-improving cycle of skill evolution during deployment. 

Rather than replacing the top-down paradigm, the bottom-up paradigm complements it by addressing its core limitations. Agents built under this framework make full use of deployment-time LLM tokens for reasoning. This allows them to adapt and refine their skills based on real-world feedback, effectively utilizing operational tokens for reasoning. Through diverse and autonomous exploration, agents are no longer static replicas of a central prototype; instead, they function as active researchers to improve staticness. As more agents are deployed, their diverse experiences accelerate the evolution of the shared skill library. The true value of agentic systems lies not in replicable  workflow design,  but in this collective, experience-driven knowledge base.

As a proof of concept, we instantiate this bottom-up philosophy through skill evolution in two open-ended game environments: Slay the Spire and Civilization V. We present the bottom-up skill evolution framework in \autoref{fig:pipeline}. Unlike traditional benchmarks~\cite{bench1,bench2,yao_second_half}, these two video games offer no explicit rewards, no predefined subgoals, and no APIs. Agents perceive the environment through visual outputs and interact via mouse operations, which is the same modality used by humans. This setup reflects the shift highlighted in The Second Half of AI~\cite{yao_second_half}: from benchmarking static tasks to evaluating agents in open-ended, stream-based environments. 

Experimental results show that the bottom-up agent can autonomously explore environments and successfully progress in both games. In Civilization V, the agent learns to build armies, attack City-States, and establish new cities. It completes 50 turns and unlocks 8 technologies under a zero-prior setting. In contrast, baseline agents often fail to make meaningful progress, frequently losing their starting settler or misdirecting key units such as the warrior. Similarly, in Slay the Spire, our agent clears 13 floors and achieves a 98.6\% execution response rate. These results highlight the superior learning capabilities of the bottom-up paradigm and its promise for scaling to complex, real-world environments. 

Our contribution is twofold: (1) we unify existing LLM-based agent frameworks under the top-down paradigm and identify three key limitations: static replication, dependence on human priors, and inefficient token usage, which constraint adaptability in open-ended settings; (2) we introduce a bottom-up paradigm based on learning from experience, offering a full formulation for agents to autonomously acquire, refine, and evolve skills through interaction.

\section{Related Works}
\textbf{Limitations of Top-Down Workflows.} The success of ChatGPT-3.5 in 2023 brought widespread attention to LLM-based agents. Yet, early agents soon encountered prompt sensitivity and hallucination issues when applied to real-world tasks. Under these constraints, the top-down paradigm was not merely a convenient design—it was an inevitable response. By decomposing complex tasks, enforcing structured workflows, and assigning agents to controlled subproblems, this approach offered a viable path to harness LLMs for reliable execution. Prominent examples include single-agent~\cite{reflexion} systems like ReAct~\cite{react} and Plan-and-Solve~\cite{plan_and_solve}, as well as multi-agent systems such as AutoGPT~\cite{autogpt}, MetaGPT~\cite{metagpt}, and ChatDev~\cite{chatdev}.

However, the top-down paradigm has reached a bottleneck due to three core limitations—staticness, prior dependency, and token overhead—as discussed in \autoref{sec:intro}.  These limitations impede its ability to adapt and scale in open-ended environments. For instance, the commercialized top-down agent Manus continues to rely heavily on human-engineered workflows and toolchains to expand its functional scope, exemplifying learning from human data~\cite{eraofexp}. Moreover, once deployed, top-down agents follow highly predictable execution paths, making them vulnerable to reverse engineering. In fact, OpenManus successfully replicated Manus’s core functionalities within three hours, underscoring the brittleness of this design philosophy.

\textbf{Possibilities Unlocked by Next-Gen LLMs.} Recent advances in LLMs~\cite{claude37,openai_gpt4o,deepseek,qwen2} provide a foundation for rethinking agent design. First, unified understanding and generation across modalities enables agents to interpret purely visual input and express complex outputs—bridging perception and action. Second, state-of-the-art models such as DeepSeek-R1~\cite{deepseek}, GPT-o-series~\cite{openai_gpt4o}, and Claude 3.5/3.7~\cite{claude37} exhibit strong reasoning capabilities, allowing agents to reflect on consequences, revise skills, and perform introspective action evaluation. Third, executable code generation~\cite{uitars,openai_operator} equips agents with the ability to synthesize and run control logic—whether for mouse/keyboard interaction in software environments or motion planning in embodied systems.

\textbf{Bottom-Up Skill Evolution.} These capabilities enable a shift in agent design—from imitating human data to emulating the human learning process itself. Human knowledge and skill acquisition typically unfold incrementally, shaped by lived experience rather than imposed top-down.  This aligns with the “era of experience” proposed by \citet{eraofexp}, which calls for a shift from curated human trajectories to learning from unstructured streams of experience.

Some early work has explored experience-driven agents. Voyager~\cite{voyager} iteratively learns tool-using skills in Minecraft, but its learning process is scaffolded by task-specific prompts and privileged APIs. Open-ended RL frameworks~\cite{vpt,voyager,nature_mine,dreamerv3} like POET~\cite{wang2019poet} and MOO~\cite{moo} encourage behavioral diversity across tasks, yet they operate without the structured reasoning capabilities of modern LLMs. In contrast, we propose a bottom-up agent paradigm that leverages LLMs’ reasoning~\cite{reasoning1,cot,reasoning2,reasoning3} and perception to acquire skills directly from raw experience—without predefined goals, tools (APIs), or strategy—enabling agents to evolve competence organically in open-ended environments.

\begin{figure}[h] 
\vspace{-1em}
	 \centering
	 \includegraphics[width = 1.0\linewidth]{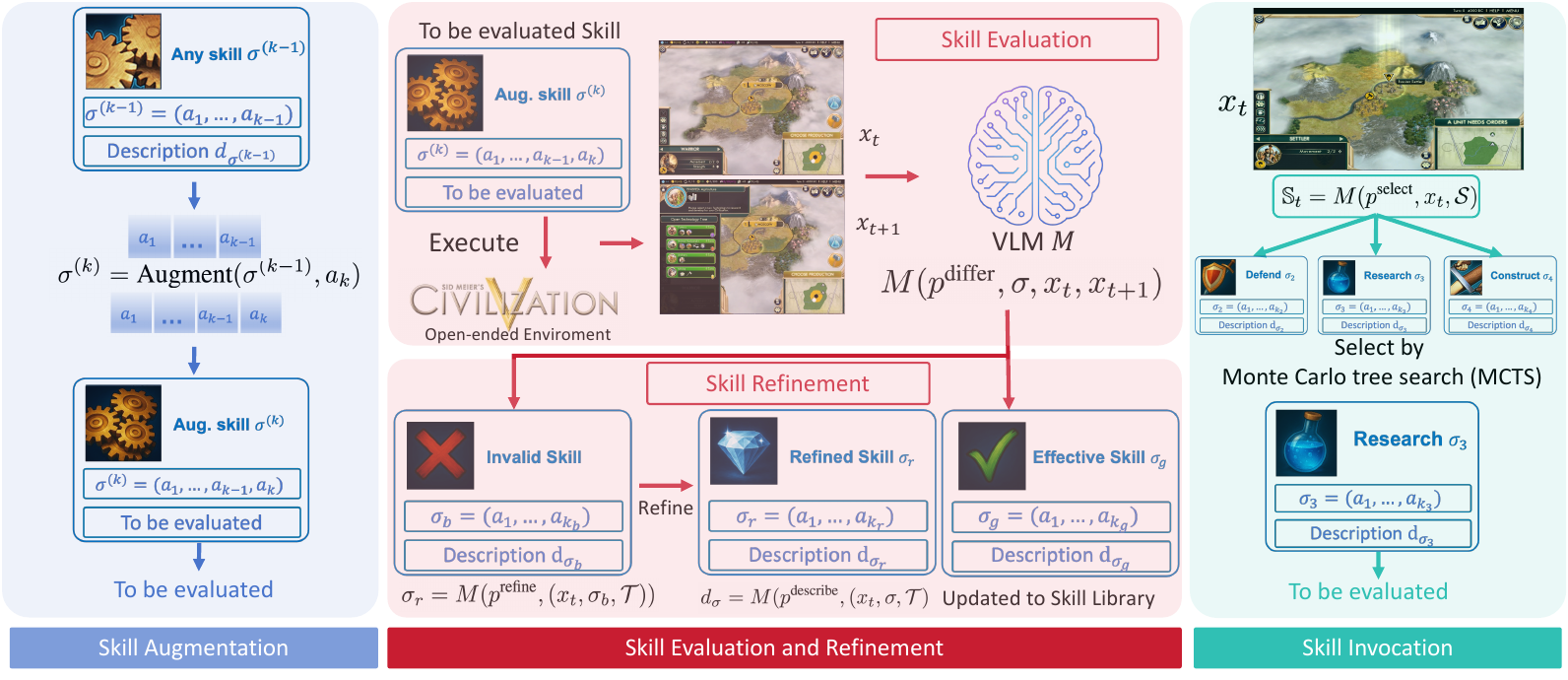}
	 \caption{\textbf{Overview of Bottom-Up Skill Evolution.} The agent begins with no predefined skills and gradually builds its library $\mathcal{S}$ through interaction. \textbf{Left:} New skills are incrementally composed by extending existing routines with atomic actions. \textbf{Middle:} Skills are evaluated by a visual-language model (VLM) comparing pre- and post-execution states; ineffective ones are refined or discarded via LLM reasoning. \textbf{Right:} At each timestep, a candidate set $\mathbb{S}_t$ is selected based on the current state $x_t$ and evaluated via Monte Carlo Tree Search (MCTS)~\cite{mcts} to choose the most promising skill. All components operate under a unified reasoning framework, without privileged APIs, allowing agents to acquire competence purely from experience.}
     \label{fig:module}
\end{figure}

\section{Methodology}

This section formalizes the bottom-up paradigm through the mechanism of skill evolution—the process by which agents construct, invoke, and refine skills through interaction. We begin by modeling the environment and defining the agent’s action and skill spaces. This formulation is grounded in a key principle: all complex skills can be composed from low-level human-like actions, such as mouse clicks and key presses. This compositionality forms the theoretical basis for bottom-up competence. The pseudocode of our bottom-up skill evolution can be found in \textcolor{myblue}{Algorithm}~\ref{algo:main}. We also illustrate the overview of our bottom-up skill evolution in \autoref{fig:module}.


\subsection{Problem Formulation}


\textbf{Notations.} We model the environment as a partially observable Markov decision process (POMDP) defined by $(\mathcal{X}, \mathcal{A}, T, \mathcal{R})$, where $\mathcal{X}$ is the observation space (visual-only) and $x_t \in \mathcal{X}$ denotes the observation at time t; $\mathcal{A}$ is the set of atomic actions (e.g., mouse clicks, drags, key presses), $T$ denotes the unknown transition dynamics, and $\mathcal{R}$ is the implicit reward signal embedded in the environment. The agent is powered by a LLM $M$, which is prompt-conditioned to serve multiple roles during interaction. Executing the LLM with a prompt $p^{(f)}$ and a current context $c_t$ yields a function-specific output: $M(p^{(f)}, c_t) \mapsto y^{(f)}_t$.

A skill is a sequence of atomic actions $\sigma = (a_1, \dots, a_k)$, where each $a_i \in \mathcal{A}$ and $k$ controls its complexity. Each skill is paired with a semantic descriptor $d_\sigma=M(p^{\text{describe}}, (x_t, \sigma, \mathcal{T}))$, which is a natural language summary of its intent. The skill library is denoted by $\mathcal{S} = \{ \sigma_1, \dots, \sigma_n \}$ and evolves over time as new skills are discovered, refined, or composed.

\textbf{Objective.} We model skill evolution as a process of continual optimization over interaction sequences. The evolution of the skill library is formulated as a population-level optimization objective:
\vspace{-0.1em}
\begin{align}
    &\max_{\mathcal{S}} \ \mathop{\mathbb{E}}\limits_{x_t \sim \mathcal{X},\ \sigma \sim \mathcal{S}} \Big[ \mathcal{R}_{\text{skill}}(\sigma, \mathcal{S}, x_t, \mathcal{T}) \Big] \quad \mbox{where} \quad \mathcal{T} = (x_{t+1}, \dots, x_{t+k}), \\
   &\mbox{and} \quad  \mathcal{R}_{\text{skill}}(\sigma, \mathcal{S}, x_t, \mathcal{T}) = R_{\text{diversity}}(\sigma, \mathcal{S}) + R_{\text{efficiency}}(\sigma) + R_{\text{semantics}}(d_\sigma, \mathcal{T}).  \label{eq:reward}
\end{align}

For any skill $\sigma = (a_1, \dots, a_k)$ invoked under state $x_t$, the agent receives an outcome trajectory $\mathcal{T}$, from which a behavioral signal is derived. Since no external reward $\mathcal{R}$ is available, the quality of a skill is assessed via an implicit $\mathcal{R}_{\text{skill}}$. 

Each term of $\mathcal{R}_{\text{skill}}$ reflects a distinct aspect of skill quality: $R_{\text{diversity}}$ encourages semantic deviation of the skill $\sigma$ from existing skills in the library $\mathcal{S}$, promoting behavioral diversity. $R_{\text{efficiency}}$ penalizes unnecessarily long or redundant action sequences, favoring concise execution. $R_{\text{semantics}}$ measures alignment between the predicted high-level effect of the skill (as described by $d_\sigma$) and the actual outcome observed in the environment.

\subsection{Skill Augmentation, Invocation, and Refinement}
In the bottom-up paradigm, agents do not begin with a fixed set of high-level behaviors, i.e., $\mathcal{S}=\emptyset$. They continuously augment, invoke, and refine skills through interaction with the environment, guided by reasoning from the LLM. 

\textbf{Skill Augmentation.} In open-ended environments with no predefined APIs and priors, most atomic actions $a \in \mathcal{A}$ are task-irrelevant and meaningless. Therefore, discovering useful skills $\sigma = (a_1, \dots, a_k)$ is through a trial-and-reasoning process: the agent explores combinations of atomic actions and reflects on their outcomes to identify meaningful behaviors.

To reduce the search space, agents incrementally construct skills by increasing skill sequence length. We use $k$ to control the length of the sequence, and it begins by evaluating single-step actions ($k = 1$), then expands to longer sequences ($k = 2, 3, \dots$), pruning unproductive branches based on behavioral feedback. Only sequences that produce recognizable effects, such as GUI transitions or task progression, are retained as meaningful candidates. The core augmentation process can thus be formalized as:
\vspace{-0.1em}
\begin{equation}
\sigma^{(k)} = \text{Augment}(\sigma^{(k-1)}, a_k), \quad \text{where } \sigma^{(k-1)} = (a_1, \dots, a_{k-1}), \ a_k \in \mathcal{A}
\end{equation}

Each new skill $\sigma^{(k)}$ is constructed by appending a random atomic action to a validated skill $\sigma^{(k-1)}$. To filter out meaningless behaviors, only sequences that trigger observable environmental changes are retained. This simple criterion ensures that the skill library preserves only potentially useful routines, enabling complex behaviors to emerge gradually from simple primitives—a core principle of bottom-up skill evolution.

\textbf{Skill Invocation.} Suppose multiple identical agents are concurrently exploring environments while sharing a common skill library $\mathcal{S}$. Given a new observation $x_t$, each agent queries the shared skill library $\mathcal{S}$ to form a candidate skill set $\mathbb{S}_t \subseteq \mathcal{S}$ using LLM reasoning:
\vspace{-0.1em}
\begin{equation}
\mathbb{S}_t = M(p^{\text{select}}, x_t, \mathcal{S})
\end{equation}

Each candidate in $\mathbb{S}_t$ represents a high-level routine potentially applicable to the current context. If no viable skill is found (i.e., $\mathbb{S}_t = \emptyset$), the agent falls back to skill augmentation in current $x_t$. However, due to environmental stochasticity and partial observability, agents avoid committing greedily. Instead, they evaluate $\mathbb{S}_t$ using Monte Carlo Tree Search (MCTS), simulating future rollouts to estimate the expected behavioral utility of each candidate. The most promising skill will be executed. 

\textbf{Skill Evaluation and Refinement.}  
Executed skills will be evaluated by the implicit reward defined in \autoref{eq:reward}. Low-scoring skills are pruned from the library after repeated poor evaluations across agents. Meanwhile, when the set of viable skills becomes sparse (e.g., $\mathbb{S}_t = \emptyset$), the agent invokes the LLM to produce a semantically refined variant:
\vspace{-0.1em}
\begin{equation}
\sigma’ = M(p^{\text{refine}}, (x_t, \sigma, \mathcal{T}))
\end{equation}
\vspace{-0.1em}
If the new candidate $\sigma’$ shows improved alignment or efficiency, it replaces the original. This two-pronged refinement, population-based pruning and LLM-guided rewriting, ensures that $\mathcal{S}$ evolves toward a more compact, reusable, and semantically consistent repertoire of skills.

\section{Instantiating Bottom-Up Agents}
To demonstrate the feasibility and generality of the bottom-up paradigm, we deploy a single-agent framework across two distinct open-ended games: Slay the Spire and Civilization V. These environments are intentionally chosen for their lack of explicit rewards, task-specific priors, and well-defined APIs, where top-down agents often struggle to scale~\cite{yao_second_half}. In contrast, bottom-up agents are naturally suited to such settings: their ability to autonomously explore, abstract, and refine skills allows them to operate in an environment-agnostic manner.

\textbf{Perception and Control Interface.} The agent perceives the environment purely through pixel-level visual observations and acts via simulated mouse inputs, closely mimicking human gameplay. At each timestep, the observation $x_t$ is obtained as a screenshot, while actions are executed as atomic control events. Crucially, the agent has no access to privileged APIs or structured game state. 

A central challenge in this setup is visual grounding, which is to interpret and act upon raw visual inputs. For instance, a click operation could theoretically target any pixel on the screen, making the action space prohibitively large. To reduce this ambiguity, we incorporate the Segment Anything Model (SAM)~\cite{sam} to dynamically identify and segment UI elements and potential interaction targets. The reconized UI elements will be updated into the skill library as well. Despite this assistance, visual grounding remains one of the most brittle components in real-world deployment of bottom-up agents, often limiting their practicality and generalization.

\textbf{Implicit Reward and Skill Refinement.}
In our setup, skill rewards are implicitly inferred from environmental feedback in two open-ended games. Specifically, we use only the semantic reward to guide skill evaluation. This is computed via the visual difference between screenshots before and after skill execution as a proxy for semantic reward, i.e., $R_{\text{semantics}} = M(p^{\text{differ}}, \sigma, x_t, x_{t+1})$. Given a new observation $x_t$, the agent does not generate skills from scratch; instead, it prompts the LLM to select a relevant subset $\mathbb{S}_t \subseteq \mathcal{S}$. These candidates are evaluated via a lightweight Monte Carlo Tree Search (MCTS)~\cite{mcts}, using implicit rewards to guide selection. Skills with consistently poor performance are refined or pruned through LLM-guided revision.

The current skills operate primarily in a record-and-replay manner: they encode fixed sequences of low-level actions. While reusable within a given environment, these routines lack the functional abstraction and parameterization needed for broader generalization. One key reason is that transforming skills into callable, modular functions typically introduces environment-specific variables—such as slot bindings to UI elements or screen coordinates, which conflicts with our goal of maintaining environment-agnostic prompting and reasoning. Nevertheless, enabling such functional abstraction remains a major opportunity for improvement, as it would allow agents to generalize more flexibly and compactly across diverse scenarios.

\textbf{Unified Deployment Across Environments.}
The same agent architecture is deployed in both Slay the Spire and Civilization V, without any environment-specific prompts, game-related rules, or prior knowledge. We expect the bottom-up agent to be truly generative to various enviroments. Therefore, all prompts used for skill generation, selection, and refinement are deliberately designed to be environment-agnostic, enabling the agent to apply the same reasoning logic across diverse settings without modification.

While the agent architecture and prompting remain identical, the skill library itself is currently environment-specific: skills acquired in one game are not directly transferable to another due to differences in visual semantics, action consequences, and UI layout. These turn-based games were chosen for the high latency of LLMs: they allow for tractable reasoning and skill reuse within each environment without real-time constraints, providing a suitable substrate for studying long-horizon skill evolution. Nonetheless, the underlying mechanism—visual grounding, compositional control, and LLM-based reasoning—remains general. As LLM capabilities continue to advance, we anticipate that this approach can naturally extend to more complex 3D environments and even embodied agents in the physical world.

\section{Experiments}
\label{sec:exp}
We evaluate our bottom-up agents in Slay the Spire and Civilization V, two structurally distinct open-ended games without task-specific priors or well-defined APIs. Our main results are presented in \autoref{tab:main_result}, demonstrating the evolution of skills in \autoref{tab:skill_evolution}. We further conduct ablation studies, with findings detailed in \autoref{tab:ablation}. These evaluations are designed to address our core question: can an agent, like a human, acquire competence entirely from scratch through experience?


\subsection{Experimental Setup}
\textbf{Environments.} We evaluate LLM-based agents in two open-ended games: Slay the Spire, a roguelike deck-builder with procedural levels, and Civilization V, a turn-based strategy game involving complex planning and resource management. Both lack explicit rewards, subgoal, APIs, or task scaffolding, where top-down agents typically falter. We use the base difficulty in Slay the Spire (Ascension 0) and the “Prince” level in Civilization V, where AI opponents gain mild advantages. Their turn-based nature accommodates LLM reasoning under current latency constraints, while real-time environments are left for future work as inference efficiency improves. 

\begin{table}[h]
\vspace{-1em}
 \setlength\tabcolsep{1.3pt} 
	 \scriptsize
     \renewcommand{\arraystretch}{1.4}
\centering
\caption{\textbf{Main results across two open-ended games.} The bottom-up agent operates under a zero-prior setting without APIs, explicit rewards, handcrafted goals, or interface bindings. As baselines struggle under these constraints, we also evaluate them with environment-specific priors (e.g., game objectives, UI control knowledge) to better highlight the potential of our bottom-up agent.}
\label{tab:main_result}
\begin{tabular}{l|cccl|cccl}
\toprule
\multirow{2}{*}{Method} & \multicolumn{4}{c|}{\textbf{Slay the Spire}} & \multicolumn{4}{c}{\textbf{Civilization V}} \\
\cmidrule(lr){2-5} \cmidrule(lr){6-9}

& \makecell{Progression $\uparrow$\\(Floors Cleared)} & \makecell{In-game $\uparrow$\\Scores} & \makecell{Execution $\uparrow$\\Responsive Rate}  & \makecell{Token Costs $\downarrow$\\(USD \$)} & \makecell{Progression $\uparrow$\\(Turns)} & \makecell{Techs $\uparrow$\\Researched} & \makecell{Execution $\uparrow$\\Responsive Rate}  & \makecell{Token Costs $\downarrow$\\(USD \$)} \\
\midrule
\multicolumn{9}{l}{\textit{Zero Priors}} \\
GPT-4o~\cite{openai_gpt4o}           &        1       &      5        &       71.4\%     &       \$ 12.05      &         6      &    0        &      36.06\%     &       \$ 7.23       \\
Claude 3.7~\cite{claude37}           &         1      &      5        &      26.09\%      &       \$ 10.13       &         0      &      0      &      15.43\%        &     \$ 10.08         \\
UITARS-1.5~\cite{uitars}              &        1       &        5      &       81.09 \%     &      \textbf{\$ 0.93}        &        0       &     0       &      61.97 \%       &     \textbf{\$ 0.95}       \\
\midrule
\multicolumn{9}{l}{\textit{With Priors}} \\
GPT-4o~\cite{openai_gpt4o}           &        8       &      46        &     51.40\%       &   \$ 9.80          &           13    &        1    &     57.24\%         &  \$ 8.94             \\
Claude 3.7~\cite{claude37}           &       1        &       5       &      79.72\%      &   \$ 12.05           &      17          &     3       &       92.91\%       &      \$ 11.45        \\
UITARS-1.5~\cite{uitars}              &        1       &      5        &     52.59\%       &   \$ 1.19           &        10       &      1      &      \textbf{93.00\%}      &      \$ 1.09         \\ 
\midrule
\makecell{\textit{Zero Priors}\\\textbf{Bottom-Up agent}}        &       \textbf{ 13}       &      \textbf{ 81}       &     \textbf{98.56\%}       &    \$ 7.14 \tablefootnote{The evaluation terminates at step 301 due to the character being defeated by the monster.}         &         \textbf{50}      &   \textbf{8}         &       92.27\%       &     \$ 6.89\tablefootnote{The evaluation terminates at step 232 due to the nation being defeated by the AI.}          \\
\bottomrule
\end{tabular}
\label{tab:baseline_comparison}
\end{table}

\textbf{Zero Prior.} Each agent starts with no subgoals, game-specific knowledge (not even the game name), or access to privileged APIs. Interaction begins from raw pixel inputs and proceeds via low-level atomic actions, which mirrors human gameplay. This zero-prior constraint is unique to our bottom-up setting; baselines are given basic priors (e.g., subgoals, game-specific knowledge) without which they fail to progress meaningfully. The prompts with zero prior of our bottom-up agent is presented in \autoref{fig:prompt}. The prompts with task-specific priors of the baselines are presented in \autoref{fig:base_prompt}. 

\textbf{LLM Backends and Deployment Protocol.} We use GPT-4o as the agent backend. Each agent runs for three episodes per environment, terminating on game completion, failure, or after a fixed action limit of 1,000 steps. A run of 1,000 steps typically takes approximately 6.5 hours. All skill acquisition, invocation, and refinement steps follow the algorithmic loop defined in Algorithm 1. Importantly, both games share an identical codebase. This unified deployment highlights the generality of the bottom-up paradigm. 

\textbf{Baselines.} Our work is the first to systematically address the problem of deploying agents in open-ended environments without APIs, subgoals, or structured observations. In such settings, most existing agent frameworks are inapplicable—they rely on predefined interfaces or task-specific prompts, and even with added priors (e.g., subgoals or game knowledge), they cannot operate effectively due to the absence of executable APIs. In practice, using these frameworks is no more effective than prompting LLMs directly. The closest comparable systems are UI agents, which interact with visual interfaces via simulated clicks and keystrokes. We therefore select GPT-4o, Claude 3.7, and the open-source UI-TARS-1.5~\cite{uitars} as representative baselines for comparison.

\textbf{Evaluation Metrics.} We assess bottom-up agents using four metrics that capture both behavioral competence and computational efficiency across environments:
(a) \textit{Progression}: measured as floors cleared in Slay the Spire and total turns played in Civilization V, reflecting sustained engagement and survival;
(b) \textit{Strategic Development}: quantified via cumulative in-game score in Slay the Spire and number of technologies researched in Civilization V, indicating planning effectiveness;
(c) \textit{Execution Responsive Rate}: the percentage of skill invocations that lead to observable changes in game state, assessing functional validity of behaviors;
(d) \textit{Token Costs}: Total LLM tokens consumed during one episode, reflecting reasoning overhead during deployment. We convert token usage into actual USD costs for better clarity.

\begin{table}[h]
\vspace{-1em}
  \setlength\tabcolsep{6pt} 
  \scriptsize
  \renewcommand{\arraystretch}{1.4}
\centering
\caption{\textbf{Skill evolution analysis} in \textit{Slay the Spire}. We track the growth of the skill library across multiple training rounds, where each round resumes from the previous one’s library. Metrics reflect the state of the skill library at the beginning of each round. Each round is configured with 100 action steps for clearer illustration.}
\label{tab:skill_evolution}
\begin{tabular}{l|cccc|cccc}
\toprule
\multirow{2}{*}{Round} & \multicolumn{4}{c|}{\textbf{Skill Library Information}} & \multicolumn{4}{c}{\textbf{Slay the Spire}} \\
\cmidrule(lr){2-5} \cmidrule(lr){6-9}
& \makecell{Library\\Size} & \makecell{Skills \\Augmented} & \makecell{Skills \\Pruned} & \makecell{Pruning\\Rate (\%)} 
& \makecell{Progression $\uparrow$\\(Floors Cleared)} & \makecell{In-game $\uparrow$\\Scores} & \makecell{Execution $\uparrow$\\Responsive Rate}  & \makecell{Token Costs $\downarrow$\\(USD \$)}  \\
\midrule
Round 0  &    0     &    60   &    1     &     1.67\%    &     6    &    36     &     93.14\%    &   \$ 2.66     \\
Round 1  &    59     &    16   &    5     &   31.25\%      &     7    &    41     &    95.37\%     &    \$ 2.28     \\
Round 2  &    70     &     26    &     0    &    0.00\%     &    8     &    53     &    96.73\%      &   \$ 2.58       \\
Round 3  &     96    &     16    &     1    &   6.25\%      &    8     &     48    &    96.58\%      &   \$ 2.41       \\
\bottomrule
\end{tabular}
\vspace{-1em}
\end{table}


\subsection{Evaluation Results}
We report the main results of experiments in \autoref{tab:main_result} and visualize the comparison of game progression in \autoref{fig:pipeline}\textcolor{myblue}{(b)}. We also provide  demo videos as supplementary to showcase our bottom-up agents successfully mastering the two games. Compared to the baseline agents both with and without environment-specific priors, our bottom-up agent demonstrates superior performance across two distinct open-ended games. In Slay the Spire, where all baselines fail to progress without priors, our agent clears 13 floors and achieves a game score of 81. We outperform all baselines, including those with access to handcrafted subgoals and interface priors. Our bottom-up agent attains a 98.56\% execution responsive rate, indicating that the discovered skills are not only meaningful but also highly functional. In Civilization V, our agent completes 50 turns and unlocks 8 technologies, showcasing effective exploration even under zero-prior constraints. While baseline performance improves marginally with added priors, they still fall short of matching our agent’s ability to sustain and adapt behaviors across timesteps.

\begin{figure}[h] 
\vspace{-.5em}
	 \centering
	 \includegraphics[width = 1.0\linewidth]{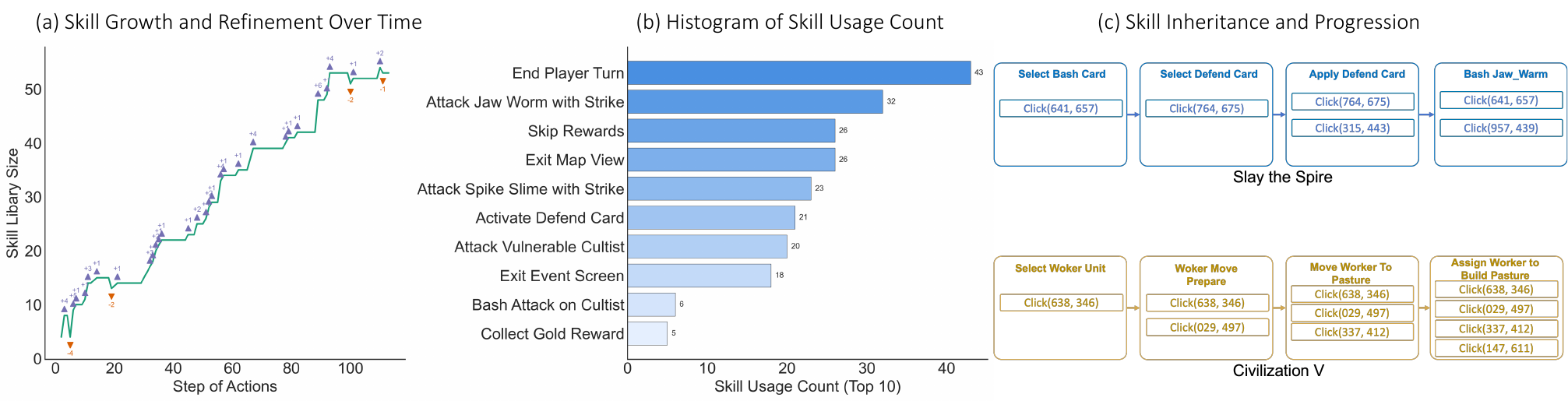}
	 \caption{\textbf{Analysis of skill evolution and reuse. (a)} Skill library size increases over time through augmentation (+) and pruning (–). \textbf{(b)}Top-10 most frequently invoked skills in Slay the Spire. \textbf{(c)} Examples of compositional skill inheritance across environments, showing how higher-level routines are built from atomic actions. }
     \label{fig:skill}
\vspace{-.5em}
\end{figure}

The primary challenge in progressing through both games lies in executing two to three precise actions in sequence. For example, in Slay the Spire, one of the most demanding skills involves dragging a card to accurately target and attack a monster. In Civilization V, the key difficulty is controlling a unit to move or attack enemy. All baseline agents are capable of interpreting visual observations, but consistently fail to execute correct actions based on that understanding. In particular, the baseline UITARS-1.5~\cite{uitars} is able to perceive game progression and articulate the next intended action. However, due to overfitting to its training data, the output action coordinates are often incorrect and misaligned with the unseen game interface.
\begin{table}[h]
\vspace{-1em}
  \setlength\tabcolsep{5.pt} 
  \scriptsize
  \renewcommand{\arraystretch}{1.4}
  \caption{\textbf{Ablation study of core components in the bottom-up agent.} We assess the contribution of three key modules: visual change filtering during skill augmentation, MCTS-based selection during skill invocation, and LLM-generated skill description during skill refinement. Fully removing the entire module halts progression entirely, so we ablate a component of each module individually.  Metrics include gameplay progression, in-game score, execution effectiveness, and token efficiency. }
\label{tab:ablation}
\begin{tabular}{l|ccc|cccc}
\toprule
\multirow{2}{*}{Setting} & \makecell{Visual Change\\Filter}  & \makecell{MCTS\\Selection} &  \makecell{Skill\\Description}
& \makecell{Progression $\uparrow$\\(Floors Cleared)} & \makecell{In-game $\uparrow$\\Scores} & \makecell{Execution $\uparrow$\\Responsive Rate} & \makecell{Token Costs $\downarrow$\\(USD \$)} \\
\midrule

w/o Visual Filter     &        & \cmark & \cmark  &   5    &   33    &    89.29\%     &   \$ 2.53     \\
w/o MCTS                & \cmark &        & \cmark &   1   &   5    &   64.52\%    &     \$ 1.48   \\
w/o Description       & \cmark & \cmark &         &    5   &    22   &    92.77\%   &      \$ 2.15  \\
Full Module (Ours)       & \cmark & \cmark & \cmark  &   6    &   36    &   93.14\%    &   \$ 2.66    \\
\bottomrule
\end{tabular}
\vspace{-1em}
\end{table}

\subsection{Ablation Studies}
We conduct ablation studies to analyze two core aspects of our framework: the effectiveness of skill evolution over time and the contribution of individual modules to overall performance.

\textbf{Skill Evolution over Rounds.}
As shown in \autoref{tab:skill_evolution}, the agent undergoes four training rounds, each consisting of 100 steps. We limit each round to 100 steps to prevent the character from being defeated by enemies—longer episodes often result in premature termination. Despite this constraint, we observe clear signs of skill evolution: the skill library grows steadily, with new skills added and non-functional ones pruned. Execution effectiveness and game score improve consistently across rounds, eventually converging in Round 4. This supports our claim that skill reuse and refinement lead to measurable behavioral improvement over time. \autoref{fig:skill} illustrates the internal dynamics of bottom-up skill evolution. More analysis of \autoref{fig:skill} can be found in \autoref{sec:more_ana}.

\begin{figure}[h] 
\vspace{-.5em}
	 \centering
	 \includegraphics[width = 1.0\linewidth]{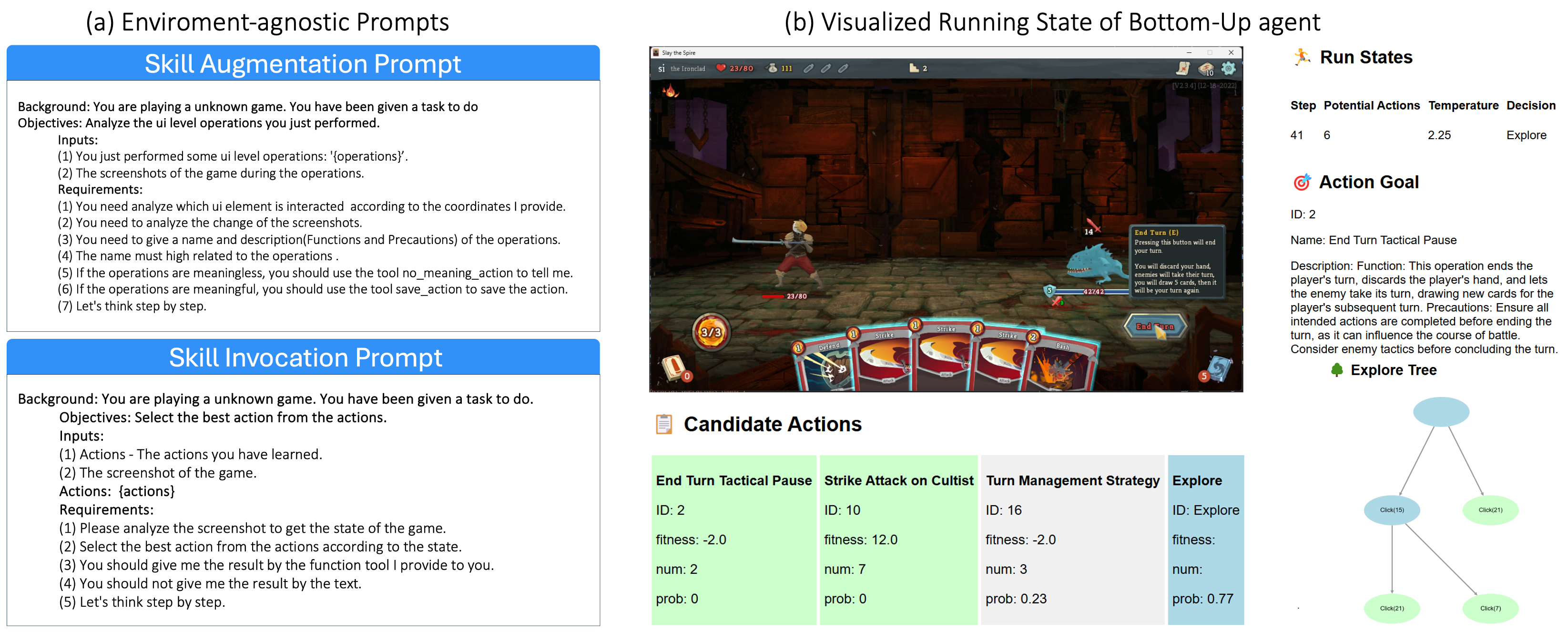}
	 \caption{\textbf{Prompting and execution visualization of the bottom-up agent. (a)}  Environment-agnostic prompts used for skill augmentation and invocation, enabling reasoning without access to game-specific APIs. \textbf{(b)} We design a GUI to visualized execution state of the agent during gameplay, showing candidate actions, selected goal, reasoning metadata and the corresponding skill plan tree. }
     \label{fig:prompt}
\vspace{-.5em}
\end{figure}

\textbf{Effect of Core Modules.}
To evaluate each module’s contribution, we ablate one component at a time: visual change filtering during augmentation, MCTS-based selection during invocation, and LLM-generated skill description during refinement. As shown in \autoref{tab:ablation}, removing any single module impairs performance across progression, in-game scores, and execution responsiveness, though to varying degrees. Notably, removing MCTS results in near-complete failure to progress, indicating its critical role in long-horizon skill planning. Disabling visual change filtering inflates token usage and reduces effective execution rate, as many invalid skills are retained. Without semantic description, the agent struggles to consolidate behaviors, leading to lower in-game scores and library reuse. 


\section{Limitations, Challenges, and Future Work}
\label{sec:limitations}
While this work offers a proof of concept for the bottom-up paradigm, building a practical and scalable agent system remains an open challenge. Our current implementation shows that learning from interaction alone is feasible, but it is not yet a comprehensive solution. Below, we outline key limitations and open problems for future work.

\textbf{Exploration Overhead.}
One current limitation of our bottom-up agent is its exploration efficiency. Since skills must be discovered and validated purely through trial-and-reasoning without task-specific priors or APIs, the agent typically requires 2-2.5 times more environment interactions to reach comparable progression levels relative to prior-assisted baselines. For example, baseline agents require approximately 6 hours to complete 1,000 interaction steps in either game, whereas our bottom-up agents take around 12 hours due to the additional reasoning and skill refinement processes involved. This additional exploration cost is a direct consequence of the zero-prior setting and the need to bootstrap useful behaviors from scratch. Future work could mitigate this overhead through more efficient skill proposal mechanisms, transfer across similar environments, or memory-based generalization from past deployments.

\textbf{Reset and Evaluation Protocols.} A fundamental challenge in open-ended environments is the lack of reliable resetting mechanisms. Unlike benchmark tasks with well-defined episode boundaries and reproducible seeds, open-ended games often lack fine-grained control over initial conditions. This introduces significant variance during evaluation and makes systematic comparisons across agents or iterations difficult. Developing standardized protocols for episode resetting and controlled skill re-evaluation will be critical for more robust benchmarking of bottom-up systems.

\textbf{Perception of Subtle Environmental Changes.} Our current reward modeling relies on perceptible visual changes (e.g., GUI transitions or progression cues) to trigger semantic credit assignment. However, many meaningful behaviors—such as defensive preparations or long-term setups—produce only subtle or delayed visual signals. The current framework may fail to detect or reward such latent strategies. A promising direction is to combine reinforcement learning techniques with implicit reward shaping, where agents learn to attribute credit over extended horizons and from nuanced patterns of environmental feedback.

\textbf{Implicit Rewards and Strategic Refinement.}
Despite these limitations, the current framework already enables agents to discover skills that effectively advance the game—e.g., clearing turns, acquiring items, or defeating enemies. However, further refinement requires transitioning from reactive behaviors to long-term, strategic play. Since no explicit reward signals exist, agents must rely on delayed environmental changes to evaluate skill quality. Introducing reinforcement learning techniques may help uncover deeper reward structures and promote strategy-level improvements, enabling skills to evolve not just in form, but in function.

\textbf{Skill Abstraction and Functionalization.}
While our agents can accumulate reusable routines via record-and-replay, these skills remain as flat action sequences rather than modular, parameterized functions. Abstracting such routines into callable, general-purpose skills is fundamentally difficult under a zero-prior setting—without predefined APIs or environment structure, it is unclear which parts of a skill should be parameterized, and even harder to infer valid parameter values from raw interaction. Without this abstraction, skill reuse remains limited to surface-level repetition rather than generalizable logic.

\textbf{Asynchronous Multi-Agent Skill Evolution.} Our framework assumes a shared but globally consistent skill library across agents. However, when multiple agents explore asynchronously, conflicts may arise: concurrent edits, incompatible refinements, or redundant additions. Future work should explore decentralized coordination mechanisms—such as eventual consistency protocols, versioned skill records, or trust-weighted refinement consensus—to ensure coherent library evolution in massively parallel deployments.

These limitations underscore that bottom-up agent design is still in its early stages. Addressing these challenges will be essential to scaling the framework from proof-of-concept to widely deployable systems.

\section{Conclusion}

In this work, we revisit LLM-based agent design by framing existing systems within a top-down paradigm, where agents act as architects to execute human-engineered goals, subtasks, and workflows. While effective in closed settings, such agents struggle to generalize beyond predefined structures. To address this, we propose a bottom-up paradigm, where agents function as explorers: acquiring, refining, and reusing skills through interaction, guided solely by implicit environmental feedback. We formalize this paradigm and implement it as a working system, demonstrating that agents can build competence autonomously from raw experience. 

Rather than replacing top-down approaches, our method complements them by enabling skill discovery in environments lacking predefined goals and APIs. In two open-ended games, our bottom-up agents exhibit emergent competence and behavioral efficiency, despite starting with no prior knowledge. Though only a proof of concept, our system points toward a promising direction for agent development. The bottom-up paradigm envisions a future where millions of agents operate across diverse environments. They share a unified and evolving skill library. The true value of agentic systems lies not in barrier-free workflow design, but in an experience-driven skill library continually honed by the collective experience of large-scale, real-world deployment.

\section*{Broader Impacts}
This work presents a bottom-up agent framework for autonomous skill acquisition through interaction, without relying on human-defined goals or APIs. While it enables scalable learning in open-ended environments and reduces dependence on manual supervision, uncontrolled deployment may lead to unintended behaviors if implicit rewards misalign with human intent. As the framework scales to real-world domains, safety, oversight, and skill traceability will become critical considerations. This study is confined to virtual environments; no real-world deployment is conducted.

{
\bibliography{neurips_2025}

\begin{thebibliography}{58}
\providecommand{\natexlab}[1]{#1}
\providecommand{\url}[1]{\texttt{#1}}
\expandafter\ifx\csname urlstyle\endcsname\relax
  \providecommand{\doi}[1]{doi: #1}\else
  \providecommand{\doi}{doi: \begingroup \urlstyle{rm}\Url}\fi

\bibitem[Achiam et~al.(2023)Achiam, Adler, Agarwal, Ahmad, Akkaya, Aleman, Almeida, Altenschmidt, Altman, Anadkat, et~al.]{gpt4}
Josh Achiam, Steven Adler, Sandhini Agarwal, Lama Ahmad, Ilge Akkaya, Florencia~Leoni Aleman, Diogo Almeida, Janko Altenschmidt, Sam Altman, Shyamal Anadkat, et~al.
\newblock Gpt-4 technical report.
\newblock \emph{arXiv preprint arXiv:2303.08774}, 2023.

\bibitem[Ahn et~al.(2022)Ahn, Brohan, Brown, Chebotar, Cortes, David, Finn, Fu, Gopalakrishnan, Hausman, Herzog, Ho, Hsu, Ibarz, Ichter, Irpan, Jang, Ruano, Jeffrey, Jesmonth, Joshi, Julian, Kalashnikov, Kuang, Lee, Levine, Lu, Luu, Parada, Pastor, Quiambao, Rao, Rettinghouse, Reyes, Sermanet, Sievers, Tan, Toshev, Vanhoucke, Xia, Xiao, Xu, Xu, Yan, and Zeng]{ahn2022icanisay}
Michael Ahn, Anthony Brohan, Noah Brown, Yevgen Chebotar, Omar Cortes, Byron David, Chelsea Finn, Chuyuan Fu, Keerthana Gopalakrishnan, Karol Hausman, Alex Herzog, Daniel Ho, Jasmine Hsu, Julian Ibarz, Brian Ichter, Alex Irpan, Eric Jang, Rosario~Jauregui Ruano, Kyle Jeffrey, Sally Jesmonth, Nikhil~J Joshi, Ryan Julian, Dmitry Kalashnikov, Yuheng Kuang, Kuang-Huei Lee, Sergey Levine, Yao Lu, Linda Luu, Carolina Parada, Peter Pastor, Jornell Quiambao, Kanishka Rao, Jarek Rettinghouse, Diego Reyes, Pierre Sermanet, Nicolas Sievers, Clayton Tan, Alexander Toshev, Vincent Vanhoucke, Fei Xia, Ted Xiao, Peng Xu, Sichun Xu, Mengyuan Yan, and Andy Zeng.
\newblock Do as i can, not as i say: Grounding language in robotic affordances, 2022.
\newblock URL \url{https://arxiv.org/abs/2204.01691}.

\bibitem[{Anthropic}(2025)]{claude37}
{Anthropic}.
\newblock Claude 3.7 sonnet and claude code.
\newblock \emph{Anthropic Blog}, February 2025.
\newblock \url{https://www.anthropic.com/news/claude-3-7-sonnet}.

\bibitem[Baker et~al.(2022)Baker, Akkaya, Zhokov, Huizinga, Tang, Ecoffet, Houghton, Sampedro, and Clune]{vpt}
Bowen Baker, Ilge Akkaya, Peter Zhokov, Joost Huizinga, Jie Tang, Adrien Ecoffet, Brandon Houghton, Raul Sampedro, and Jeff Clune.
\newblock Video pretraining (vpt): Learning to act by watching unlabeled online videos.
\newblock \emph{Advances in Neural Information Processing Systems}, 35:\penalty0 24639--24654, 2022.

\bibitem[Bonatti et~al.(2024)Bonatti, Zhao, Bonacci, Dupont, Abdali, Li, Lu, Wagle, Koishida, Bucker, et~al.]{bench2}
Rogerio Bonatti, Dan Zhao, Francesco Bonacci, Dillon Dupont, Sara Abdali, Yinheng Li, Yadong Lu, Justin Wagle, Kazuhito Koishida, Arthur Bucker, et~al.
\newblock Windows agent arena: Evaluating multi-modal os agents at scale.
\newblock \emph{arXiv preprint arXiv:2409.08264}, 2024.

\bibitem[Brohan et~al.(2023{\natexlab{a}})Brohan, Brown, Carbajal, Chebotar, Chen, Choromanski, Ding, Driess, Dubey, Finn, Florence, Fu, Arenas, Gopalakrishnan, Han, Hausman, Herzog, Hsu, Ichter, Irpan, Joshi, Julian, Kalashnikov, Kuang, Leal, Lee, Lee, Levine, Lu, Michalewski, Mordatch, Pertsch, Rao, Reymann, Ryoo, Salazar, Sanketi, Sermanet, Singh, Singh, Soricut, Tran, Vanhoucke, Vuong, Wahid, Welker, Wohlhart, Wu, Xia, Xiao, Xu, Xu, Yu, and Zitkovich]{brohan2023rt2visionlanguageactionmodelstransfer}
Anthony Brohan, Noah Brown, Justice Carbajal, Yevgen Chebotar, Xi~Chen, Krzysztof Choromanski, Tianli Ding, Danny Driess, Avinava Dubey, Chelsea Finn, Pete Florence, Chuyuan Fu, Montse~Gonzalez Arenas, Keerthana Gopalakrishnan, Kehang Han, Karol Hausman, Alexander Herzog, Jasmine Hsu, Brian Ichter, Alex Irpan, Nikhil Joshi, Ryan Julian, Dmitry Kalashnikov, Yuheng Kuang, Isabel Leal, Lisa Lee, Tsang-Wei~Edward Lee, Sergey Levine, Yao Lu, Henryk Michalewski, Igor Mordatch, Karl Pertsch, Kanishka Rao, Krista Reymann, Michael Ryoo, Grecia Salazar, Pannag Sanketi, Pierre Sermanet, Jaspiar Singh, Anikait Singh, Radu Soricut, Huong Tran, Vincent Vanhoucke, Quan Vuong, Ayzaan Wahid, Stefan Welker, Paul Wohlhart, Jialin Wu, Fei Xia, Ted Xiao, Peng Xu, Sichun Xu, Tianhe Yu, and Brianna Zitkovich.
\newblock Rt-2: Vision-language-action models transfer web knowledge to robotic control, 2023{\natexlab{a}}.
\newblock URL \url{https://arxiv.org/abs/2307.15818}.

\bibitem[Brohan et~al.(2023{\natexlab{b}})Brohan, Brown, Carbajal, Chebotar, Dabis, Finn, Gopalakrishnan, Hausman, Herzog, Hsu, Ibarz, Ichter, Irpan, Jackson, Jesmonth, Joshi, Julian, Kalashnikov, Kuang, Leal, Lee, Levine, Lu, Malla, Manjunath, Mordatch, Nachum, Parada, Peralta, Perez, Pertsch, Quiambao, Rao, Ryoo, Salazar, Sanketi, Sayed, Singh, Sontakke, Stone, Tan, Tran, Vanhoucke, Vega, Vuong, Xia, Xiao, Xu, Xu, Yu, and Zitkovich]{brohan2023rt1roboticstransformerrealworld}
Anthony Brohan, Noah Brown, Justice Carbajal, Yevgen Chebotar, Joseph Dabis, Chelsea Finn, Keerthana Gopalakrishnan, Karol Hausman, Alex Herzog, Jasmine Hsu, Julian Ibarz, Brian Ichter, Alex Irpan, Tomas Jackson, Sally Jesmonth, Nikhil~J Joshi, Ryan Julian, Dmitry Kalashnikov, Yuheng Kuang, Isabel Leal, Kuang-Huei Lee, Sergey Levine, Yao Lu, Utsav Malla, Deeksha Manjunath, Igor Mordatch, Ofir Nachum, Carolina Parada, Jodilyn Peralta, Emily Perez, Karl Pertsch, Jornell Quiambao, Kanishka Rao, Michael Ryoo, Grecia Salazar, Pannag Sanketi, Kevin Sayed, Jaspiar Singh, Sumedh Sontakke, Austin Stone, Clayton Tan, Huong Tran, Vincent Vanhoucke, Steve Vega, Quan Vuong, Fei Xia, Ted Xiao, Peng Xu, Sichun Xu, Tianhe Yu, and Brianna Zitkovich.
\newblock Rt-1: Robotics transformer for real-world control at scale, 2023{\natexlab{b}}.
\newblock URL \url{https://arxiv.org/abs/2212.06817}.

\bibitem[{Butterfly Effect AI}(2025)]{manus}
{Butterfly Effect AI}.
\newblock Manus ai: General ai agent, March 2025.
\newblock URL \url{https://manus.im/}.
\newblock Accessed: 2025-05-14.

\bibitem[Cabannes et~al.(2024)Cabannes, Arnal, Bouaziz, Yang, Charton, and Kempe]{reasoning3}
Vivien Cabannes, Charles Arnal, Wassim Bouaziz, Xingyu Yang, Francois Charton, and Julia Kempe.
\newblock Iteration head: A mechanistic study of chain-of-thought.
\newblock \emph{Advances in Neural Information Processing Systems}, 37:\penalty0 109101--109122, 2024.

\bibitem[Cemri et~al.(2025)Cemri, Pan, Yang, Agrawal, Chopra, Tiwari, Keutzer, Parameswaran, Klein, Ramchandran, et~al.]{mas_fail}
Mert Cemri, Melissa~Z Pan, Shuyi Yang, Lakshya~A Agrawal, Bhavya Chopra, Rishabh Tiwari, Kurt Keutzer, Aditya Parameswaran, Dan Klein, Kannan Ramchandran, et~al.
\newblock Why do multi-agent llm systems fail?
\newblock \emph{arXiv preprint arXiv:2503.13657}, 2025.

\bibitem[Chen et~al.(2025)Chen, Wu, Liu, Pan, Liu, Xie, Yu, and Ruan]{janus}
Xiaokang Chen, Zhiyu Wu, Xingchao Liu, Zizheng Pan, Wen Liu, Zhenda Xie, Xingkai Yu, and Chong Ruan.
\newblock Janus-pro: Unified multimodal understanding and generation with data and model scaling.
\newblock \emph{arXiv preprint arXiv:2501.17811}, 2025.

\bibitem[Espeholt et~al.(2018)Espeholt, Soyer, Munos, Simonyan, Mnih, Ward, Doron, Firoiu, Harley, Dunning, Legg, and Kavukcuoglu]{espeholt2018impalascalabledistributeddeeprl}
Lasse Espeholt, Hubert Soyer, Remi Munos, Karen Simonyan, Volodymir Mnih, Tom Ward, Yotam Doron, Vlad Firoiu, Tim Harley, Iain Dunning, Shane Legg, and Koray Kavukcuoglu.
\newblock Impala: Scalable distributed deep-rl with importance weighted actor-learner architectures, 2018.
\newblock URL \url{https://arxiv.org/abs/1802.01561}.

\bibitem[Eysenbach et~al.(2018)Eysenbach, Gupta, Ibarz, and Levine]{eysenbach2018diversityneedlearningskills}
Benjamin Eysenbach, Abhishek Gupta, Julian Ibarz, and Sergey Levine.
\newblock Diversity is all you need: Learning skills without a reward function, 2018.
\newblock URL \url{https://arxiv.org/abs/1802.06070}.

\bibitem[Guo et~al.(2025)Guo, Yang, Zhang, Song, Zhang, Xu, Zhu, Ma, Wang, Bi, et~al.]{deepseek}
Daya Guo, Dejian Yang, Haowei Zhang, Junxiao Song, Ruoyu Zhang, Runxin Xu, Qihao Zhu, Shirong Ma, Peiyi Wang, Xiao Bi, et~al.
\newblock Deepseek-r1: Incentivizing reasoning capability in llms via reinforcement learning.
\newblock \emph{arXiv preprint arXiv:2501.12948}, 2025.

\bibitem[Hafner et~al.(2023)Hafner, Pasukonis, Ba, and Lillicrap]{dreamerv3}
Danijar Hafner, Jurgis Pasukonis, Jimmy Ba, and Timothy Lillicrap.
\newblock Mastering diverse domains through world models.
\newblock \emph{arXiv preprint arXiv:2301.04104}, 2023.

\bibitem[Hafner et~al.(2025)Hafner, Pasukonis, Ba, and Lillicrap]{nature_mine}
Danijar Hafner, Jurgis Pasukonis, Jimmy Ba, and Timothy Lillicrap.
\newblock Mastering diverse control tasks through world models.
\newblock \emph{Nature}, pages 1--7, 2025.

\bibitem[He et~al.(2024)He, Yao, Ma, Yu, Dai, Zhang, Lan, and Yu]{bench3}
Hongliang He, Wenlin Yao, Kaixin Ma, Wenhao Yu, Yong Dai, Hongming Zhang, Zhenzhong Lan, and Dong Yu.
\newblock Webvoyager: Building an end-to-end web agent with large multimodal models.
\newblock \emph{arXiv preprint arXiv:2401.13919}, 2024.

\bibitem[Hong et~al.(2023)Hong, Zheng, Chen, Cheng, Wang, Zhang, Wang, Yau, Lin, Zhou, et~al.]{metagpt}
Sirui Hong, Xiawu Zheng, Jonathan Chen, Yuheng Cheng, Jinlin Wang, Ceyao Zhang, Zili Wang, Steven Ka~Shing Yau, Zijuan Lin, Liyang Zhou, et~al.
\newblock Metagpt: Meta programming for multi-agent collaborative framework.
\newblock \emph{arXiv preprint arXiv:2308.00352}, 3\penalty0 (4):\penalty0 6, 2023.

\bibitem[Hu et~al.(2025)Hu, Guo, Wang, Chen, Wang, Zhang, Sreenath, Lu, and Chen]{hu2025videopredictionpolicygeneralist}
Yucheng Hu, Yanjiang Guo, Pengchao Wang, Xiaoyu Chen, Yen-Jen Wang, Jianke Zhang, Koushil Sreenath, Chaochao Lu, and Jianyu Chen.
\newblock Video prediction policy: A generalist robot policy with predictive visual representations, 2025.
\newblock URL \url{https://arxiv.org/abs/2412.14803}.

\bibitem[Jiang et~al.(2023)Jiang, Gupta, Zhang, Wang, Dou, Chen, Fei-Fei, Anandkumar, Zhu, and Fan]{jiang2023vimageneralrobotmanipulation}
Yunfan Jiang, Agrim Gupta, Zichen Zhang, Guanzhi Wang, Yongqiang Dou, Yanjun Chen, Li~Fei-Fei, Anima Anandkumar, Yuke Zhu, and Linxi Fan.
\newblock Vima: General robot manipulation with multimodal prompts, 2023.
\newblock URL \url{https://arxiv.org/abs/2210.03094}.

\bibitem[Kirillov et~al.(2023)Kirillov, Mintun, Ravi, Mao, Rolland, Gustafson, Xiao, Whitehead, Berg, Lo, et~al.]{sam}
Alexander Kirillov, Eric Mintun, Nikhila Ravi, Hanzi Mao, Chloe Rolland, Laura Gustafson, Tete Xiao, Spencer Whitehead, Alexander~C Berg, Wan-Yen Lo, et~al.
\newblock Segment anything.
\newblock In \emph{Proceedings of the IEEE/CVF international conference on computer vision}, pages 4015--4026, 2023.

\bibitem[Leibo et~al.(2021)Leibo, Duéñez-Guzmán, Vezhnevets, Agapiou, Sunehag, Koster, Matyas, Beattie, Mordatch, and Graepel]{leibo2021scalableevaluationmultiagentreinforcement}
Joel~Z. Leibo, Edgar Duéñez-Guzmán, Alexander~Sasha Vezhnevets, John~P. Agapiou, Peter Sunehag, Raphael Koster, Jayd Matyas, Charles Beattie, Igor Mordatch, and Thore Graepel.
\newblock Scalable evaluation of multi-agent reinforcement learning with melting pot, 2021.
\newblock URL \url{https://arxiv.org/abs/2107.06857}.

\bibitem[Li et~al.(2023)Li, Hammoud, Itani, Khizbullin, and Ghanem]{li2023camelcommunicativeagentsmind}
Guohao Li, Hasan Abed Al~Kader Hammoud, Hani Itani, Dmitrii Khizbullin, and Bernard Ghanem.
\newblock Camel: Communicative agents for "mind" exploration of large language model society, 2023.
\newblock URL \url{https://arxiv.org/abs/2303.17760}.

\bibitem[Liang et~al.(2023)Liang, Huang, Xia, Xu, Hausman, Ichter, Florence, and Zeng]{liang2023codepolicieslanguagemodel}
Jacky Liang, Wenlong Huang, Fei Xia, Peng Xu, Karol Hausman, Brian Ichter, Pete Florence, and Andy Zeng.
\newblock Code as policies: Language model programs for embodied control, 2023.
\newblock URL \url{https://arxiv.org/abs/2209.07753}.

\bibitem[Liu et~al.(2023)Liu, Yu, Zhang, Xu, Lei, Lai, Gu, Ding, Men, Yang, et~al.]{bench1}
Xiao Liu, Hao Yu, Hanchen Zhang, Yifan Xu, Xuanyu Lei, Hanyu Lai, Yu~Gu, Hangliang Ding, Kaiwen Men, Kejuan Yang, et~al.
\newblock Agentbench: Evaluating llms as agents.
\newblock \emph{arXiv preprint arXiv:2308.03688}, 2023.

\bibitem[Mnih et~al.(2015)Mnih, Kavukcuoglu, Silver, Rusu, Veness, Bellemare, Graves, Riedmiller, Fidjeland, Ostrovski, Petersen, Beattie, Sadik, Antonoglou, King, Kumaran, Wierstra, Legg, and Hassabis]{mnih2015humanlevel}
Volodymyr Mnih, Koray Kavukcuoglu, David Silver, Andrei~A Rusu, Joel Veness, Marc~G Bellemare, Alex Graves, Martin Riedmiller, Andreas~K Fidjeland, Georg Ostrovski, Stig Petersen, Charles Beattie, Amir Sadik, Ioannis Antonoglou, Helen King, Dharshan Kumaran, Daan Wierstra, Shane Legg, and Demis Hassabis.
\newblock Human-level control through deep reinforcement learning.
\newblock \emph{Nature}, 518\penalty0 (7540):\penalty0 529--533, 2015.

\bibitem[Mnih et~al.(2016)Mnih, Badia, Mirza, Graves, Lillicrap, Harley, Silver, and Kavukcuoglu]{mnih2016asynchronousmethodsdeepreinforcement}
Volodymyr Mnih, Adrià~Puigdomènech Badia, Mehdi Mirza, Alex Graves, Timothy~P. Lillicrap, Tim Harley, David Silver, and Koray Kavukcuoglu.
\newblock Asynchronous methods for deep reinforcement learning, 2016.
\newblock URL \url{https://arxiv.org/abs/1602.01783}.

\bibitem[{OpenAI}(2024)]{openai_gpt4o}
{OpenAI}.
\newblock Gpt-4o: Openai's multimodal flagship model.
\newblock \emph{OpenAI Blog}, May 2024.
\newblock \url{https://openai.com/index/hello-gpt-4o/}.

\bibitem[{OpenAI}(2025)]{openai_operator}
{OpenAI}.
\newblock Introducing operator: Openai's autonomous web agent.
\newblock \emph{OpenAI Blog}, January 2025.
\newblock \url{https://openai.com/index/introducing-operator/}.

\bibitem[Park et~al.(2023)Park, O'Brien, Cai, Morris, Liang, and Bernstein]{generative_agent}
Joon~Sung Park, Joseph O'Brien, Carrie~Jun Cai, Meredith~Ringel Morris, Percy Liang, and Michael~S Bernstein.
\newblock Generative agents: Interactive simulacra of human behavior.
\newblock In \emph{Proceedings of the 36th annual acm symposium on user interface software and technology}, pages 1--22, 2023.

\bibitem[Qian et~al.(2023)Qian, Liu, Liu, Chen, Dang, Li, Yang, Chen, Su, Cong, et~al.]{chatdev}
Chen Qian, Wei Liu, Hongzhang Liu, Nuo Chen, Yufan Dang, Jiahao Li, Cheng Yang, Weize Chen, Yusheng Su, Xin Cong, et~al.
\newblock Chatdev: Communicative agents for software development.
\newblock \emph{arXiv preprint arXiv:2307.07924}, 2023.

\bibitem[Qin et~al.(2025)Qin, Ye, Fang, Wang, Liang, Tian, Zhang, Li, Li, Huang, et~al.]{uitars}
Yujia Qin, Yining Ye, Junjie Fang, Haoming Wang, Shihao Liang, Shizuo Tian, Junda Zhang, Jiahao Li, Yunxin Li, Shijue Huang, et~al.
\newblock Ui-tars: Pioneering automated gui interaction with native agents.
\newblock \emph{arXiv preprint arXiv:2501.12326}, 2025.

\bibitem[Rawles et~al.(2024)Rawles, Clinckemaillie, Chang, Waltz, Lau, Fair, Li, Bishop, Li, Campbell-Ajala, et~al.]{bench4}
Christopher Rawles, Sarah Clinckemaillie, Yifan Chang, Jonathan Waltz, Gabrielle Lau, Marybeth Fair, Alice Li, William Bishop, Wei Li, Folawiyo Campbell-Ajala, et~al.
\newblock Androidworld: A dynamic benchmarking environment for autonomous agents.
\newblock \emph{arXiv preprint arXiv:2405.14573}, 2024.

\bibitem[Schulman et~al.(2017)Schulman, Wolski, Dhariwal, Radford, and Klimov]{schulman2017proximalpolicyoptimizationalgorithms}
John Schulman, Filip Wolski, Prafulla Dhariwal, Alec Radford, and Oleg Klimov.
\newblock Proximal policy optimization algorithms, 2017.
\newblock URL \url{https://arxiv.org/abs/1707.06347}.

\bibitem[Shaikh et~al.(2022)Shaikh, Zhang, Held, Bernstein, and Yang]{reasoning2}
Omar Shaikh, Hongxin Zhang, William Held, Michael Bernstein, and Diyi Yang.
\newblock On second thought, let's not think step by step! bias and toxicity in zero-shot reasoning.
\newblock \emph{arXiv preprint arXiv:2212.08061}, 2022.

\bibitem[Shinn et~al.(2023)Shinn, Cassano, Gopinath, Narasimhan, and Yao]{reflexion}
Noah Shinn, Federico Cassano, Ashwin Gopinath, Karthik Narasimhan, and Shunyu Yao.
\newblock Reflexion: Language agents with verbal reinforcement learning.
\newblock \emph{Advances in Neural Information Processing Systems}, 36:\penalty0 8634--8652, 2023.

\bibitem[Silver and Sutton(2025)]{eraofexp}
David Silver and Richard~S Sutton.
\newblock Welcome to the era of experience.
\newblock 2025.

\bibitem[Stone et~al.(2023)Stone, Xiao, Lu, Gopalakrishnan, Lee, Vuong, Wohlhart, Kirmani, Zitkovich, Xia, et~al.]{moo}
Austin Stone, Ted Xiao, Yao Lu, Keerthana Gopalakrishnan, Kuang-Huei Lee, Quan Vuong, Paul Wohlhart, Sean Kirmani, Brianna Zitkovich, Fei Xia, et~al.
\newblock Open-world object manipulation using pre-trained vision-language models.
\newblock \emph{arXiv preprint arXiv:2303.00905}, 2023.

\bibitem[Su et~al.(2025)Su, Sun, Yoon, Yin, Yu, and Arık]{su2025learnbyinteractdatacentricframeworkselfadaptive}
Hongjin Su, Ruoxi Sun, Jinsung Yoon, Pengcheng Yin, Tao Yu, and Sercan~Ö. Arık.
\newblock Learn-by-interact: A data-centric framework for self-adaptive agents in realistic environments, 2025.
\newblock URL \url{https://arxiv.org/abs/2501.10893}.

\bibitem[{\'S}wiechowski et~al.(2023){\'S}wiechowski, Godlewski, Sawicki, and Ma{\'n}dziuk]{mcts}
Maciej {\'S}wiechowski, Konrad Godlewski, Bartosz Sawicki, and Jacek Ma{\'n}dziuk.
\newblock Monte carlo tree search: A review of recent modifications and applications.
\newblock \emph{Artificial Intelligence Review}, 56\penalty0 (3):\penalty0 2497--2562, 2023.

\bibitem[Team et~al.(2021)Team, Stooke, Mahajan, Barros, Deck, Bauer, Sygnowski, Trebacz, Jaderberg, Mathieu, McAleese, Bradley-Schmieg, Wong, Porcel, Raileanu, Hughes-Fitt, Dalibard, and Czarnecki]{openendedlearningteam2021openendedlearningleadsgenerally}
Open Ended~Learning Team, Adam Stooke, Anuj Mahajan, Catarina Barros, Charlie Deck, Jakob Bauer, Jakub Sygnowski, Maja Trebacz, Max Jaderberg, Michael Mathieu, Nat McAleese, Nathalie Bradley-Schmieg, Nathaniel Wong, Nicolas Porcel, Roberta Raileanu, Steph Hughes-Fitt, Valentin Dalibard, and Wojciech~Marian Czarnecki.
\newblock Open-ended learning leads to generally capable agents, 2021.
\newblock URL \url{https://arxiv.org/abs/2107.12808}.

\bibitem[Vezhnevets et~al.(2017)Vezhnevets, Osindero, Schaul, Heess, Jaderberg, Silver, and Kavukcuoglu]{vezhnevets2017feudalnetworkshierarchicalreinforcement}
Alexander~Sasha Vezhnevets, Simon Osindero, Tom Schaul, Nicolas Heess, Max Jaderberg, David Silver, and Koray Kavukcuoglu.
\newblock Feudal networks for hierarchical reinforcement learning, 2017.
\newblock URL \url{https://arxiv.org/abs/1703.01161}.

\bibitem[Wang et~al.(2023{\natexlab{a}})Wang, Xie, Jiang, Mandlekar, Xiao, Zhu, Fan, and Anandkumar]{voyager}
Guanzhi Wang, Yuqi Xie, Yunfan Jiang, Ajay Mandlekar, Chaowei Xiao, Yuke Zhu, Linxi Fan, and Anima Anandkumar.
\newblock Voyager: An open-ended embodied agent with large language models.
\newblock \emph{arXiv preprint arXiv:2305.16291}, 2023{\natexlab{a}}.

\bibitem[Wang et~al.(2023{\natexlab{b}})Wang, Xu, Lan, Hu, Lan, Lee, and Lim]{plan_and_solve}
Lei Wang, Wanyu Xu, Yihuai Lan, Zhiqiang Hu, Yunshi Lan, Roy Ka-Wei Lee, and Ee-Peng Lim.
\newblock Plan-and-solve prompting: Improving zero-shot chain-of-thought reasoning by large language models.
\newblock \emph{arXiv preprint arXiv:2305.04091}, 2023{\natexlab{b}}.

\bibitem[Wang et~al.(2019)Wang, Lehman, Clune, and Stanley]{wang2019poet}
Rui Wang, Joel Lehman, Jeff Clune, and Kenneth~O. Stanley.
\newblock Paired open-ended trailblazer (poet): Endlessly generating increasingly complex and diverse learning environments and their solutions, 2019.
\newblock URL \url{https://arxiv.org/abs/1901.01753}.

\bibitem[Wang et~al.(2025)Wang, Wang, Liu, Ding, Zhang, Liu, and Zhang]{token_eff1}
Zhexuan Wang, Yutong Wang, Xuebo Liu, Liang Ding, Miao Zhang, Jie Liu, and Min Zhang.
\newblock Agentdropout: Dynamic agent elimination for token-efficient and high-performance llm-based multi-agent collaboration.
\newblock \emph{arXiv preprint arXiv:2503.18891}, 2025.

\bibitem[Wei et~al.(2022{\natexlab{a}})Wei, Tay, Bommasani, Raffel, Zoph, Borgeaud, Yogatama, Bosma, Zhou, Metzler, et~al.]{reasoning1}
Jason Wei, Yi~Tay, Rishi Bommasani, Colin Raffel, Barret Zoph, Sebastian Borgeaud, Dani Yogatama, Maarten Bosma, Denny Zhou, Donald Metzler, et~al.
\newblock Emergent abilities of large language models.
\newblock \emph{arXiv preprint arXiv:2206.07682}, 2022{\natexlab{a}}.

\bibitem[Wei et~al.(2022{\natexlab{b}})Wei, Wang, Schuurmans, Bosma, Xia, Chi, Le, Zhou, et~al.]{cot}
Jason Wei, Xuezhi Wang, Dale Schuurmans, Maarten Bosma, Fei Xia, Ed~Chi, Quoc~V Le, Denny Zhou, et~al.
\newblock Chain-of-thought prompting elicits reasoning in large language models.
\newblock \emph{Advances in neural information processing systems}, 35:\penalty0 24824--24837, 2022{\natexlab{b}}.

\bibitem[Wu et~al.(2023)Wu, Bansal, Zhang, Wu, Li, Zhu, Jiang, Zhang, Zhang, Liu, Awadallah, White, Burger, and Wang]{wu2023autogenenablingnextgenllm}
Qingyun Wu, Gagan Bansal, Jieyu Zhang, Yiran Wu, Beibin Li, Erkang Zhu, Li~Jiang, Xiaoyun Zhang, Shaokun Zhang, Jiale Liu, Ahmed~Hassan Awadallah, Ryen~W White, Doug Burger, and Chi Wang.
\newblock Autogen: Enabling next-gen llm applications via multi-agent conversation, 2023.
\newblock URL \url{https://arxiv.org/abs/2308.08155}.

\bibitem[Xie et~al.(2024)Xie, Zhang, Chen, Li, Zhao, Cao, Hua, Cheng, Shin, Lei, et~al.]{osworld}
Tianbao Xie, Danyang Zhang, Jixuan Chen, Xiaochuan Li, Siheng Zhao, Ruisheng Cao, Toh~J Hua, Zhoujun Cheng, Dongchan Shin, Fangyu Lei, et~al.
\newblock Osworld: Benchmarking multimodal agents for open-ended tasks in real computer environments.
\newblock \emph{Advances in Neural Information Processing Systems}, 37:\penalty0 52040--52094, 2024.

\bibitem[Yang et~al.(2024)Yang, Yang, Zhang, Hui, Zheng, Yu, Li, Liu, Huang, Wei, et~al.]{qwen2}
An~Yang, Baosong Yang, Beichen Zhang, Binyuan Hui, Bo~Zheng, Bowen Yu, Chengyuan Li, Dayiheng Liu, Fei Huang, Haoran Wei, et~al.
\newblock Qwen2. 5 technical report.
\newblock \emph{arXiv preprint arXiv:2412.15115}, 2024.

\bibitem[Yang et~al.(2023{\natexlab{a}})Yang, Yue, and He]{autogpt}
Hui Yang, Sifu Yue, and Yunzhong He.
\newblock Auto-gpt for online decision making: Benchmarks and additional opinions.
\newblock \emph{arXiv preprint arXiv:2306.02224}, 2023{\natexlab{a}}.

\bibitem[Yang et~al.(2023{\natexlab{b}})Yang, Yang, Lu, Zhou, and Li]{skill1}
Mingyu Yang, Yaodong Yang, Zhenbo Lu, Wengang Zhou, and Houqiang Li.
\newblock Hierarchical multi-agent skill discovery.
\newblock \emph{Advances in Neural Information Processing Systems}, 36:\penalty0 61759--61776, 2023{\natexlab{b}}.

\bibitem[Yao(2025)]{yao_second_half}
Shunyu Yao.
\newblock The second half.
\newblock \emph{ysymyth.github.io}, April 2025.
\newblock \url{https://ysymyth.github.io/The-Second-Half/}.

\bibitem[Yao et~al.(2023)Yao, Zhao, Yu, Du, Shafran, Narasimhan, and Cao]{react}
Shunyu Yao, Jeffrey Zhao, Dian Yu, Nan Du, Izhak Shafran, Karthik Narasimhan, and Yuan Cao.
\newblock React: Synergizing reasoning and acting in language models.
\newblock In \emph{International Conference on Learning Representations (ICLR)}, 2023.

\bibitem[Zeng et~al.(2025)Zeng, Huang, Jiang, Liu, Jin, Tiana, Li, and Xu]{token_eff2}
Yuting Zeng, Weizhe Huang, Lei Jiang, Tongxuan Liu, Xitai Jin, Chen~Tianying Tiana, Jing Li, and Xiaohua Xu.
\newblock S $^2$-mad: Breaking the token barrier to enhance multi-agent debate efficiency.
\newblock \emph{arXiv preprint arXiv:2502.04790}, 2025.

\bibitem[Zheng et~al.(2017)Zheng, Yang, Cai, Zhang, Wang, and Yu]{zheng2017magentmanyagentreinforcementlearning}
Lianmin Zheng, Jiacheng Yang, Han Cai, Weinan Zhang, Jun Wang, and Yong Yu.
\newblock Magent: A many-agent reinforcement learning platform for artificial collective intelligence, 2017.
\newblock URL \url{https://arxiv.org/abs/1712.00600}.

\bibitem[Zhou et~al.(2024)Zhou, Yang, Lin, Bai, Zhou, Wang, Levine, and Li]{skill2}
Yifei Zhou, Qianlan Yang, Kaixiang Lin, Min Bai, Xiong Zhou, Yu-Xiong Wang, Sergey Levine, and Erran Li.
\newblock Proposer-agent-evaluator (pae): Autonomous skill discovery for foundation model internet agents.
\newblock \emph{arXiv preprint arXiv:2412.13194}, 2024.

\end{thebibliography}
}

\clearpage
\appendix


\begin{algorithm}[H]
\label{algo:main}
\caption{Bottom-Up Skill Evolution}
\KwIn{Environment $(\mathcal{X}, \mathcal{A}, T, \mathcal{R})$, initialized skill library $\mathcal{S} = \emptyset$, LLM model $M$}
\ForEach{agent}{
    \While{episode not terminated}{
        Observe $x_t \in \mathcal{X}$\;
        Sample candidate skills $\mathbb{S}_t = M(p^{\text{select}}, x_t, \mathcal{S})$\;
        \eIf{$\mathbb{S}_t = \emptyset$}{
            \For{$k = 1$ \KwTo $k_{\max}$}{
                Generate $s^{(k)} = \text{Augment}(s^{(k-1)}, a_k)$\;
                \If{effect$(s^{(k)})$ is recognizable}{
                    Generate $d_\sigma = M(p^{\text{describe}}, (x_t, s^{(k)}, \mathcal{T}))$\;
                    Update $\mathcal{S} \leftarrow \mathcal{S} \cup \{ (s^{(k)}, d_\sigma) \}$\;
                    \textbf{break}\;
                }
            }
        }{
            Evaluate $\mathbb{S}_t$ with MCTS\;
            Execute the best skill $\sigma^*$ from MCTS\;
        }
        Observe trajectory $\mathcal{T}$ and compute $R_{\text{semantics}}$\;
        \If{$R_{\text{semantics}}$ low across agents}{
            Remove $\sigma$ from $\mathcal{S}$\;
        }
        \If{skill count $< \text{threshold}$}{
            Refine skill: $\sigma' = M(p^{\text{refine}}, (x_t, \sigma, \mathcal{T}))$\;
            Replace $\sigma$ with $\sigma'$ if improvement observed\;
        }
    }
}
\end{algorithm}


\section{More Related Works}
\textbf{Learning Agents in Traditional Reinforcement Learning.} Deep reinforcement learning has produced agents capable of learning complex skills from low-level inputs through trial-and-error based on predefined rewards. Notably, DQN demonstrated that a convolutional agent could achieve human-level play on multiple Atari games, with the game score as the sole reward signal \citep{mnih2015humanlevel}. Asynchronous methods like A3C allowed actor-critic agents to surpass Atari benchmarks \citep{mnih2016asynchronousmethodsdeepreinforcement}, and PPO further refined policy-gradient updates for improved reliability \citep{schulman2017proximalpolicyoptimizationalgorithms}. Scalable architectures such as IMPALA extended RL to hundreds of tasks, enabling positive knowledge transfer in a single agent trained on 3D and Atari games \citep{espeholt2018impalascalabledistributeddeeprl}. These works highlight the power of trial-and-error learning in well-defined environments. However, they struggle with transferability across environments, as they rely on task-specific reward functions and cannot easily generalize beyond the training scenario.

Classic hierarchical RL frameworks like FeUdal Networks (FuN) introduce skill abstraction by separating a “manager” that sets goals from a “worker” that executes primitive actions, yet still require predefined task hierarchies and tuning \citep{vezhnevets2017feudalnetworkshierarchicalreinforcement}. Unsupervised skill discovery methods like DIAYN maximize diversity through self-discovered behaviors without explicit rewards  \citep{eysenbach2018diversityneedlearningskills}. The bottom-up agent framework extends these ideas by accumulating behaviors into a global knowledge base, where skills are continuously refined and shared across tasks. This method not only addresses skill transferability but also reduces reliance on task-specific rewards, supporting cross-task skill reuse by design.

\textbf{Open-Ended and Multi-Agent Learning Systems.} Open-ended learning frameworks have shown the potential of dynamically generating new tasks and environments for agents, leading to more complex and adaptive behaviors over time. POET generates evolving obstacles and reuses agents across tasks, enabling continual progress in new environments \citep{wang2019poet}. Similarly, DeepMind’s XLand creates procedurally generated 3D worlds, where agents learn without human intervention and exhibit general behaviors like experimentation and innovation \citep{openendedlearningteam2021openendedlearningleadsgenerally}. Multi-agent systems, such as Melting Pot, extend these ideas by simulating complex environments with multiple interacting agents, allowing for emergent behaviors to evolve over time \citep{leibo2021scalableevaluationmultiagentreinforcement}. However, these systems often rely on centralized coordination, role assignment, and predefined communication protocols, limiting their flexibility and adaptability to new, unexpected tasks.

\begin{figure}[h] 
	 \centering
	 \includegraphics[width = 1.0\linewidth]{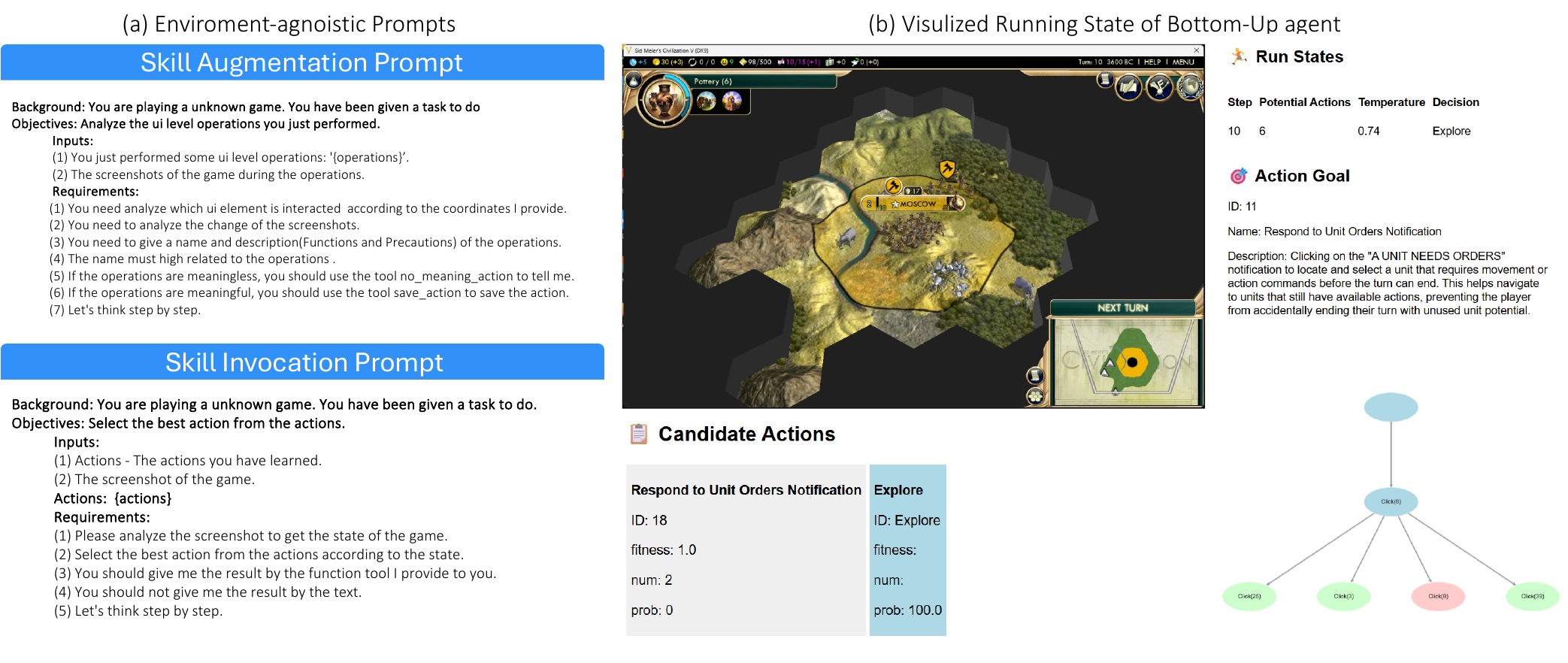}
	 \caption{\textbf{Prompting and execution visualization of the bottom-up agent. (a)}  Environment-agnostic prompts used for skill augmentation and invocation, enabling reasoning without access to game-specific APIs. We use the same codebase and prompts in both Slay the Spire and Civilizaion V. \textbf{(b)} The GUI to visualized execution state of the agent  playing Civilization V. }
     \label{fig:pro2}
\end{figure}

Many multi-agent RL platforms~\cite{skill1}, like MAgent still assume a fixed scenario or symmetric roles for agents\citep{zheng2017magentmanyagentreinforcementlearning}. Recent multi-agent coordination frameworks for LLMs take an even more structured approach: frameworks like AutoGen and CAMEL define specialized agent roles and have them converse to solve tasks\citep{wu2023autogenenablingnextgenllm,li2023camelcommunicativeagentsmind}. In these cases, the multi-agent interaction protocol or hierarchy is designed a priori—whether it’s a communication channel, a population update rule, or an explicit role assignment. Rather than relying on fixed roles and local synchronization, bottom-up agent framework shares learned skills asynchronously in a global repository. This decentralized, open-ended approach allows agents to operate independently, continuously exploring and learning, while benefiting from the discoveries of others without predefined roles, providing greater flexibility and scalability.
\begin{figure}[h] 
	 \centering
	 \includegraphics[width = 1.0\linewidth]{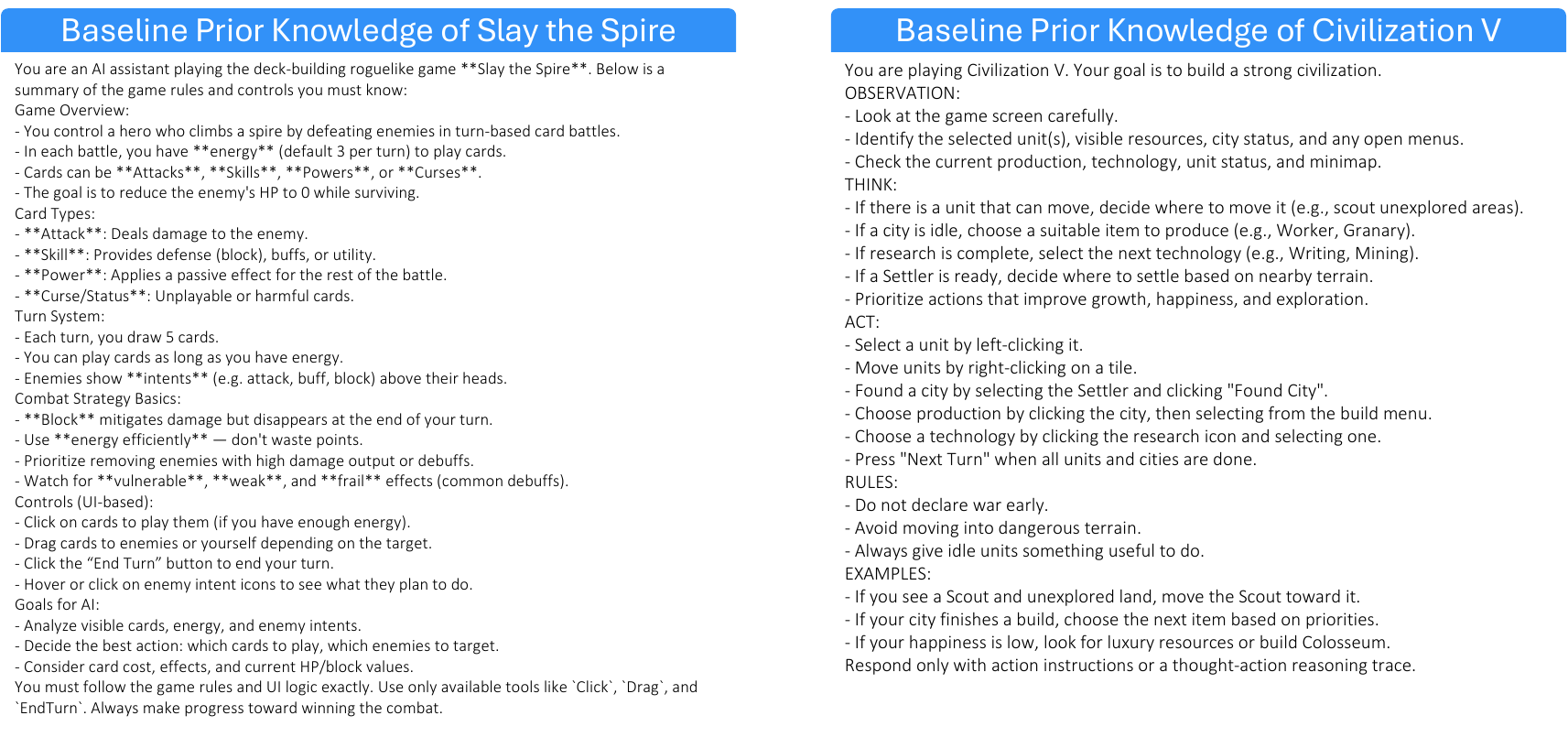}
	 \caption{\textbf{Baseline prior knowledge provided to top-down agents.}  To enable task execution in Slay the Spire (left) and Civilization V (right), baseline agents are given structured game-specific summaries detailing rules, objectives, and UI controls. These priors are necessary for baseline agents to function in the absence of direct API access.}
     \label{fig:base_prompt}
\end{figure}
\vspace{-1em}

\textbf{Embodied and Vision-Language-Action Agents.} Vision-language-action (VLA) systems, such as RT-1 and RT-2, have achieved remarkable results by combining large-scale pretraining with task-specific learning to perform complex robotic actions based on visual inputs and language commands. RT-1 uses a large dataset of human demonstrations to learn multi-task policies, and RT-2 extends this by incorporating vision-language pretraining from internet-scale data, enabling zero-shot generalization and task completion beyond prior experience \citep{brohan2023rt1roboticstransformerrealworld,brohan2023rt2visionlanguageactionmodelstransfer}. Similarly, PaLM-SayCan integrates an LLM as a high-level planner that decomposes user instructions into feasible actions, while Code-as-Policies generates executable code for robots to follow complex instructions \citep{ahn2022icanisay,liang2023codepolicieslanguagemodel}. More recent approaches, such as VIMA, train transformer-based models on multimodal inputs (images and text) to perform manipulation tasks with minimal supervision, while VPP uses video prediction models to understand physical dynamics for multi-task robotic control \citep{jiang2023vimageneralrobotmanipulation,hu2025videopredictionpolicygeneralist,skill2}. These systems excel at following structured commands but heavily depend on human-engineered demonstrations, predefined skills, and large datasets.

\begin{figure}[h] 
	 \centering
	 \includegraphics[width = 1.0\linewidth]{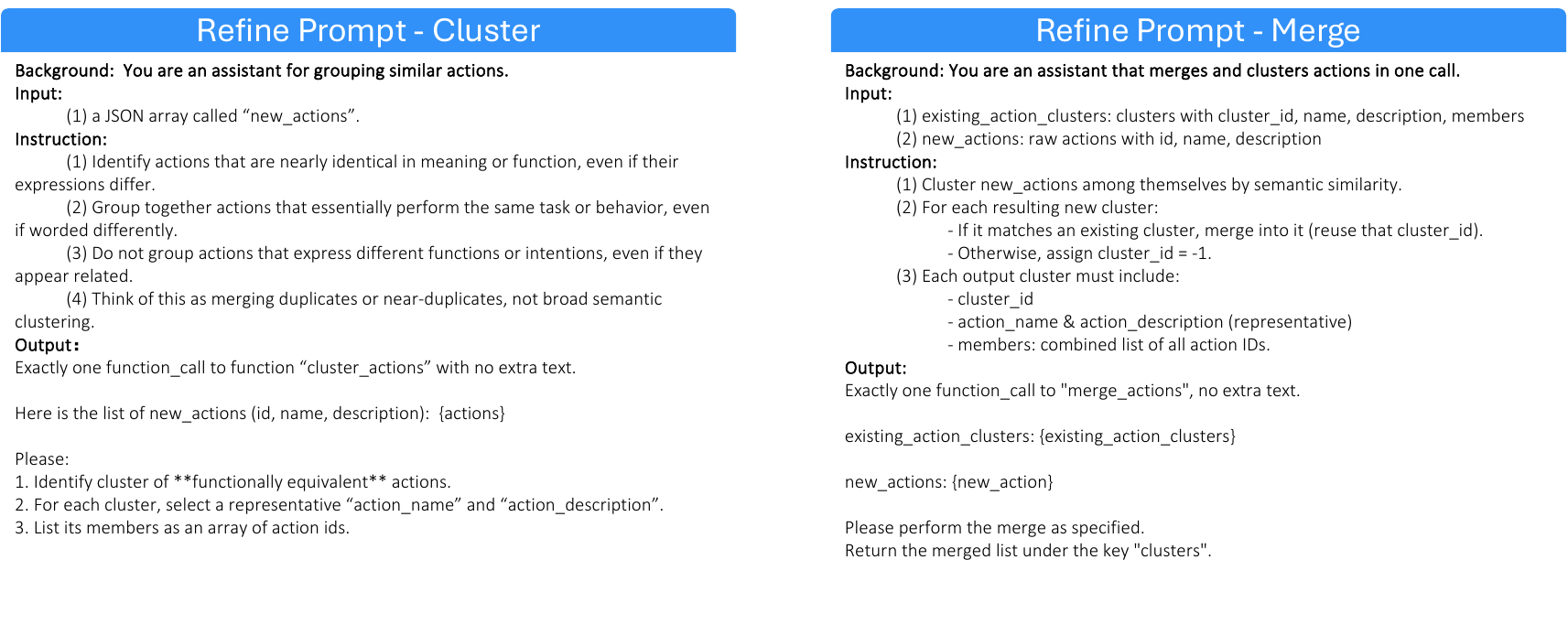}
	 \caption{\textbf{Prompts for skill augmentation via clustering and merging.}  The left prompt guides the agent to group functionally equivalent actions into clusters; the right prompt extends this by merging with existing clusters. These prompts enable semantic consolidation of skills during augmentation, reducing redundancy and promoting generalization.}
     \label{fig:refine_prompt}
\end{figure}
\vspace{-1em}

The bottom-up framework diverges from these VLA models by focusing on autonomous skill discovery through trial-and-error learning, using visual feedback as the primary source of guidance. While VLA systems leverage pre-existing knowledge or external instructions to guide behavior, our agents develop competencies by interacting with their environment directly, refining behaviors over time. Although our agents incorporate LLM APIs for reasoning and planning, they do not rely on explicit task prompts or large pre-existing datasets for skill acquisition. Instead, they abstract successful behaviors through their own experiences, enabling the emergence of transferable skills without the need for extensive human input.

Learn-by-Interact~\cite{su2025learnbyinteractdatacentricframeworkselfadaptive} synthesizes task-aligned data in realistic UI environments using documentation and backward construction. Our work, by contrast, adopts a task-free, bottom-up approach where agents acquire and refine skills purely through interaction. While their targets instruction-following via data synthesis, ours enables open-ended skill evolution without priors—making the two approaches complementary in advancing general-purpose agents.

\section{More Experimental Results}
\subsection{More analysis}
\label{sec:more_ana}
We analyze the details of \autoref{fig:skill} here. As shown in (a), the skill library gradually expands through iterative augmentation and pruning, with semantically redundant or ineffective skills removed over time. Panel (b) highlights emergent reuse, where certain high-utility skills are invoked frequently, while the majority of skills are seldom to be invoked. Panel (c) shows how low-level atomic actions are composed into higher-level routines through inheritance, forming modular, environment-specific skills in both Slay the Spire and Civilization V.
\subsection{Prompts Demonstratetion}
We demonstrate more prompts we used for baselines and our bottom-up agents. \autoref{fig:base_prompt} shows the task-specific prompts for baselines, which encode structured prior knowledge about each game. These include explicit rules, action types, UI conventions, and strategic heuristics required for baseline agents to function in the absence of APIs. In contrast, our bottom-up agents cannot access any prior knowledge to verify the environment’s rules, semantics, or action space. As shown in \autoref{fig:prompt}, they rely on general-purpose prompts to analyze, abstract, and organize skills purely from visual interaction. These prompts guide the agent to identify functionally equivalent actions, cluster them into semantically meaningful groups, and iteratively refine the skill library over time—enabling autonomous skill evolution without environment-specific guidance. We use the prompts shown in \autoref{fig:refine_prompt} to cluster and merge skills that are augmented during exploration. As the bottom-up agent interacts with the environment, it accumulates a large number of low-level skills composed of atomic UI actions. Many of these skills are functionally redundant or differ only in superficial details such as coordinates or UI layout. To manage this growing skill library and promote generalization, we periodically invoke LLM-based clustering through structured prompts.

\clearpage

\end{document}